\documentclass{article}

\usepackage{microtype}
\usepackage{graphicx}
\usepackage{subcaption}
\usepackage{booktabs} 

\usepackage{xcolor}
\usepackage{tcolorbox}
\usepackage{bm}
\usepackage{enumitem}

\definecolor{tokenGray}{HTML}{F0F0F0}

\newcommand{\token}[1]{%
  \tcbox[
    on line,
    boxsep=0pt,
    left=2pt,
    right=2pt,
    top=1pt,
    bottom=1pt,
    colback=tokenGray,
    colframe=tokenGray,
    rounded corners, arc=2pt,
    fontupper={\ttfamily}
  ]{#1}%
}

\usepackage{hyperref}

\usepackage[accepted]{icml2026}

\usepackage{amsmath}
\usepackage{amssymb}
\usepackage{mathtools}
\usepackage{amsthm}

\usepackage[capitalize,noabbrev]{cleveref}

\theoremstyle{plain}

\theoremstyle{definition}

\theoremstyle{remark}

\usepackage[textsize=tiny]{todonotes}

\newcommand{\camready}[1]{#1}                    

\newcommand{\given}{\,|\,}

\icmltitlerunning{Breaking the Simplification Bottleneck in Amortized Neural Symbolic Regression}

\begin{document}

\twocolumn[
  \icmltitle{Breaking the Simplification Bottleneck in Amortized Neural Symbolic Regression}



  \icmlsetsymbol{equal}{*}

  \begin{icmlauthorlist}
      \icmlauthor{Paul Saegert}{heidelberg}
      \icmlauthor{Ullrich Köthe}{heidelberg}
  \end{icmlauthorlist}

  \icmlaffiliation{heidelberg}{Computer Vision and Learning Lab, Heidelberg University, Germany}

  \icmlcorrespondingauthor{Paul Saegert}{bc226@uni-heidelberg.de}
  \icmlcorrespondingauthor{Ullrich Köthe}{ullrich.koethe@iwr.uni-heidelberg.de}

  \icmlkeywords{Machine Learning, ICML, Symbolic Regression, Transformer, Simulation Based Inference}

  \vskip 0.3in
]





\printAffiliationsAndNotice{}  

\begin{abstract}
  Symbolic regression (SR) aims to discover interpretable analytical expressions that accurately describe observed data.
  Amortized SR promises to be much more efficient than the predominant genetic programming SR methods, but currently struggles to scale to realistic scientific complexity.
  We find that a key obstacle is the lack of a fast reduction of equivalent expressions to a concise normalized form.
  Amortized SR has addressed this with general-purpose Computer Algebra Systems (CAS) like SymPy, but the high computational cost severely limits training and inference speed.
  We propose \textbf{SimpliPy}, a rule-based simplification engine achieving a $100$-fold speed-up over SymPy at comparable quality.
  This enables substantial improvements in amortized SR, including scalability to much larger training sets, more efficient use of the per-expression token budget, and systematic training set decontamination with respect to equivalent test expressions.
  We demonstrate these advantages in our \textbf{Flash-ANSR} framework, which achieves much better accuracy than amortized baselines (NeSymReS, E2E) on the FastSRB benchmark.
  Moreover, it performs on par with state-of-the-art direct optimization (PySR) while recovering more concise rather than more complex expressions with increasing inference budget. 
\end{abstract}

\section{Introduction}

Symbolic regression (SR) occupies a unique position in scientific machine learning by enabling the discovery of interpretable, closed-form laws from observational data \citep{Schmidt2009DistillingFN}.
Unlike standard deep learning, SR seeks the analytical expressions that govern the data-generating process.
Traditionally, this is framed as a combinatorial optimization problem solved via genetic programming (GP) \citep{Koza1994GeneticPA, cranmer2023interpretablemachinelearningscience}.
While GP remains the gold standard for precision, it treats every dataset as a \emph{tabula rasa} search instance, failing to transfer structural knowledge between tasks.

Pioneered by \citet{lample2019deeplearningsymbolicmathematics, biggio2021neural}, amortized SR solves this by learning the posterior $p(\text{expression} \given \text{data})$ over millions of examples. It aims to shift the computational burden to a one-time pre-training phase, \camready{encoding the prior of a simulation-based-inference simulator (SBI; \citealp{Cranmer_2020}) into the model}.
Meanwhile, neural scaling laws dictate that robust generalization requires data that is vast, diverse \cite{kaplan2020scalinglawsneurallanguage}, and high-quality \cite{lample2019deeplearningsymbolicmathematics, lee2022deduplicatingtrainingdatamakes, gunasekar2023textbooksneed}.
This presents a fundamental challenge: randomly generated mathematical expressions are rife with redundancies (e.g., $x+x$ vs. $2x$) that must be \emph{simplified} to ensure high-quality, normalized training targets.

Current state-of-the-art approaches face a dilemma regarding this simplification step.
Standard Computer Algebra Systems (CAS) like SymPy \citep{sympy} rely on heavy object-oriented parsing and tree traversals.
When integrated into training loops \citep{bendinelli2023controllableneuralsymbolicregression,yu2025symbolicregressionmdlformerguidedsearch}, this creates a \emph{simplification bottleneck}, where data generation becomes orders of magnitude slower than gradient updates.
To maintain feasible training times, many compromise on the amount or diversity of the training data, or abandon simplification entirely \citep{kamienny2022endtoendsymbolicregressiontransformers}.

\begin{figure}[t]
  \begin{center}
    \centerline{\includegraphics[width=0.75\columnwidth,trim=77mm 170mm 35mm 11mm,clip]{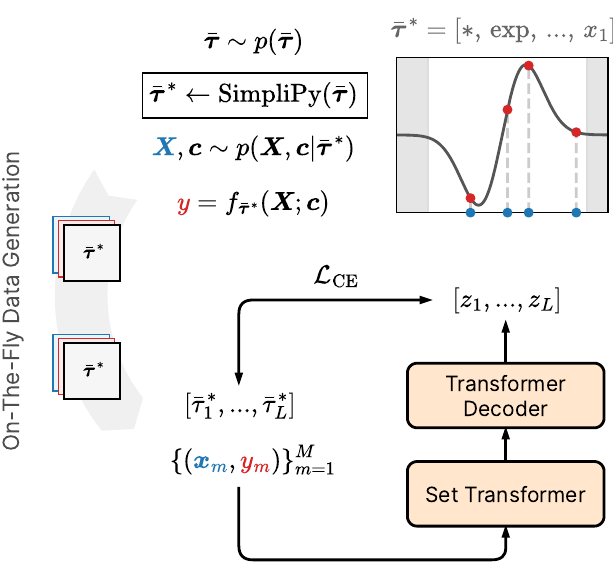}}
    \caption{
    The \textsc{Flash-ANSR} training pipeline.
    Following the established standard encoder-decoder paradigm, our framework integrates \textbf{SimpliPy} (top center) into the loop for synchronous simplification of on-the-fly generated training expressions.}
    \label{fig:flash-ansr-training}
  \end{center}
  \vspace{-25pt}
\end{figure}

We introduce \textsc{SimpliPy}\footnote{\url{https://github.com/psaegert/simplipy}}, an engine that reduces symbolic simplification to fast pattern matching, achieving speedups of up to $100\times$ over SymPy at comparable quality.
This breakthrough enables \textsc{Flash-ANSR}\footnote{\url{https://github.com/psaegert/flash-ansr}} (Figure \ref{fig:flash-ansr-training}), a framework for Transformer-based training on a continuously generated stream of high-quality expressions.
Freed from the common CAS bottleneck, \textsc{Flash-ANSR} can be scaled to learn to approximate the posterior over a much broader distribution of mathematical functions and data.

\textbf{Contributions:}
\vspace{-6pt}
\begin{enumerate}
    \item We identify the simplification bottleneck as a key obstacle to scaling the training and inference of amortized SR and introduce \textsc{SimpliPy}, a hash-based simplification engine to resolve it.
    \item We present \textsc{Flash-ANSR}, which leverages \textsc{SimpliPy} to train on 512M on-the-fly generated and simplified data-expression pairs, scaling to higher-dimensional inputs and broader operator sets than prior work.
    \item We demonstrate that \textsc{Flash-ANSR} dominates the inference-time-recovery-rate Pareto frontier against both static (NeSymReS) and unsimplified (E2E) baselines, and matches state-of-the-art GP methods (PySR) with superior inference-time-parsimony scaling.
    \item Addressing the pervasive lack of rigor regarding data leakage and evaluation standards in the field, we implement strict \textit{symbolic and numeric decontamination} of training data, ensuring our results reflect actual generalization.
    We follow a rigorous \textit{test-time compute} evaluation protocol, exposing the inference time budget trade-offs often obscured in prior literature.
\end{enumerate}

\paragraph{Conflict of Interest Disclosure.}
The authors declare no conflicts of interest.

\section{Problem Formulation}
\label{sec:problem_formulation}

\camready{
Given a dataset $\mathcal{D} = \{(\mathbf{x}_m \in \mathbb{R}^D, y_m \in \mathbb{R})\}_{m=1}^{M}$, symbolic regression (SR) seeks an analytical function $f_{\bm{\tau}}(\,\cdot\,;\mathbf{c}) \colon \mathbb{R}^D \to \mathbb{R}$ that explains the data, defined by a mathematical expression $\bm{\tau} = [\tau_1, \ldots, \tau_L]$ represented as a sequence containing input variables $\mathcal{V} = \{x_1, \ldots, x_D\}$, operators $\mathcal{O}$ (e.g.\ $\{+,*,\sin,\dots\}$), and constants $\mathbf{c} \in \mathbb{R}^K$.
The space of expressions $\mathcal{E}$ is combinatorially large, and contains many functionally distinct expressions, as well as many syntactic rewrites of functionally equivalent expressions.


}

\section{Related Work}
\label{sec:related_work}

Symbolic regression (SR) has traditionally been formulated as a combinatorial optimization problem.
While genetic programming (GP) approaches like PySR \citep{cranmer2023interpretablemachinelearningscience} and Operon \citep{operon} remain the gold standard for precision, they treat every task as a new search instance.
Amortized methods are instead framed as simulation-based inference \cite{Cranmer_2020}: they exploit a massive synthetic training corpus to learn a parameterized approximation of the posterior distribution $p(\bm{\tau} \given \mathcal{D})$ of explanatory expressions $\bm{\tau}$ for given observed data $\mathcal{D}$.
However, existing approaches are compromised by the trade-off between expression quality (i.e. simplification) and data diversity and scale.

\subsection{Static Corpora and Finite Support}
To ensure high-quality training targets, many amortized methods rely on SymPy \cite{sympy} for simplification.
Its computational cost, however, leads many to perform data generation offline, resulting in fixed, pre-computed datasets on the order of 100M expressions \cite{biggio2021neural, symformer}.
These expensive and limited training sets ultimately force a trade-off between coverage, diversity, and dimensionality: to maintain a representative density of samples over the prior, training is restricted to limited operator sets and low-dimensional settings ($D \leq 3$).
Furthermore, these static datasets are confined to iterating over a fixed subset of the prior, fundamentally restricting the diversity of functional forms the model is exposed to during training compared to on-the-fly expression generation.

\subsection{Lack of Normalization and Syntactic Redundancy}
To escape the limitations of expensive static datasets, recent works have moved toward on-the-fly procedural generation.
However, to maintain training throughput, several methods abandon simplification as a consequence, and train on vast streams of unsimplified expressions \cite{kamienny2022endtoendsymbolicregressiontransformers, Li_2025}.
This forces the model to approximate a redundant one-to-many mapping during training where a single underlying function has infinite valid representations (e.g., $x$, $x+0$, $1 \cdot x$) which dilutes the probability mass across syntactically distinct but semantically identical targets.
This learned \textit{syntactic redundancy} severely hampers inference efficiency.
The generative search dissipates its budget producing syntactic rewrites rather than distinct functional hypotheses, and, since they are not filtered, the constants have to be redundantly re-optimized for each variant.

\subsection{The SymPy Bottleneck in Dynamic Generation}
A third category of methods attempts to combine on-the-fly generation with CAS-based simplification.
NSRwH \citep{bendinelli2023controllableneuralsymbolicregression} and MDLformer \citep{yu2025symbolicregressionmdlformerguidedsearch} apply SymPy within the training loop to simplify generated skeletons.
This raises the data quality but introduces a severe computational bottleneck for large scale training.
The overhead of object-oriented parsing and tree traversals in SymPy restricts the throughput and maximum complexity of expressions that can be generated in real-time.
Consequently, these methods are often limited to lower-dimensional problems.

\textsc{Flash-ANSR} resolves this dilemma via \textsc{SimpliPy}, enabling the high-throughput generation of simplified expressions with high dimensionality and broad operator support, significantly exceeding the scale of prior dynamic methods.

\subsection{Methodological Deficiencies in Evaluation}
We identify a pervasive lack of rigor in SR concerning data handling and evaluation.
Many works report point-estimates without confidence intervals \citep{bendinelli2023controllableneuralsymbolicregression} or use lenient success thresholds ($R^2 > 0.9$) that mask failures \citep{d2023odeformer}.
Analyses often ignore inference time budgets, comparing methods with vastly different compute budgets \citep{valipour2021symbolicgpt}.
Most alarmingly, we find that with the partial exception of \citet{biggio2021neural}, virtually no prior work performs rigorous data decontamination.
This risks performance overestimation \cite{carlini2023quantifying} and violates fundamental evaluation principles \cite{hastie2009elements}.
Our work establishes a strict evaluation protocol: we demand machine-precision recovery ($\text{FVU} < 10^{-7}$), analyze the \textit{test-time compute} Pareto frontier with respect to a diverse set of metrics, and perform rigorous and conservative decontamination to prevent test set leakage.

\section{Method}
\label{sec:method}

\camready{We frame SR as a set-to-sequence task in which a transformer encoder-decoder model learns to approximate the posterior distribution $p(\bm{\tau} \given \mathcal{D})$ over expressions (i.e. token sequences in prefix notation) $\bm{\tau}$ given datasets $\mathcal{D}$.
}
Our framework comprises three components: (1) \textsc{SimpliPy}, a \camready{rule-based} engine that reduces algebraic simplification to \camready{pattern matching} and cancellations; (2) a synchronous, high-throughput data-generation and training pipeline; and (3) a specialized encoder-decoder Transformer.

\subsection{SimpliPy: Amortized Symbolic Simplification}
\label{sec:simplipy}

To enable high-throughput generation of concise expressions, we introduce \textsc{SimpliPy}\camready{, a length-reducing rule-based term rewriting system (TRS) $\mathcal{R}$ paired with a multiplicity-based cancellation procedure for additive and multiplicative subtrees.
Training only exposes the model to \emph{normalized} forms, allowing its capacity to be spent on functionally distinct hypotheses rather than on rediscovering algebraic identities.
We use the term \emph{normalized form} for the shortest representative of an expression's algebraic equivalence class under $\mathcal{R}$; among equally-short irreducible forms, we select the one with the fewest constants $n_c(\bm{\tau})$.
Among the equivalence class $[\bm{\tau}]_\mathcal{R}$ of $\bm{\tau}$, we select
\begin{equation*}
\bm{\tau}^* = \operatorname{argmin}_{\bm{\tau}' \in [\bm{\tau}]_\mathcal{R}}\bigl(|\bm{\tau}'|,\ n_c(\bm{\tau}')\bigr).
\end{equation*}
For example, $x_1 + c$ is the normalized form of the class $\{x_1 + \sin(c),\,\, 2x_1 + c_1 + c_2 - x_1,\,\, \dots\}$.
}
Standard computer algebra solves this problem from first principles, but this is overkill for the kind of expressions encountered in SR training and therefore unnecessarily slow.
Instead, we apply an \textit{amortization} strategy to the simplification process itself and determine normalized forms for (sub)expressions with up to $D=4$ variables and $L_\text{max}=7$ symbols offline and up-front\camready{. Each metavariable $\square_i$ in a rule's LHS binds to an arbitrary subtree at match time, extending the coverage of these length-bounded patterns to subtrees of arbitrary depth at runtime}.
This one-time investment allows us to reduce runtime simplification to fast table lookups.

\textsc{SimpliPy} operates directly on the tokenized prefix sequence produced by the generator and consumed by the model, allowing us to bypass additional parsing overhead caused by conversions to object-oriented structures.

\textbf{Phase 1: Amortized Rule Discovery (Offline).}
\camready{We construct $\mathcal{R}$ by stratified enumeration and numeric equivalence testing of patterns and candidate replacements in order of increasing length, similar in spirit to Kruskal's algorithm \cite{Kruskal} for minimum spanning forests:
each pattern is processed shortest-first and a new rule is added only when $\textsc{SimpliPy}_\mathcal{R}$ cannot already reduce that pattern.
To permit embarrassingly parallel discovery, $\mathcal{R}$ is held fixed within each length stratum at the cost of a fraction of redundant rules.
A formal algorithm is given in Appendix \ref{app:algorithms}.
}

\camready{Every accepted rule $\bm{\tau} \to \bm{\tau}'$ satisfies (i) $\mathrm{Vars}(\bm{\tau}') \subseteq \mathrm{Vars}(\bm{\tau})$ (no new variables are introduced) and (ii) $|\bm{\tau}'| < |\bm{\tau}|$ (strict length reduction).
The discovery loop additionally rejects any rule whose replacement uses a metavariable more often than its LHS, since such rules would not strictly reduce length in general once the metavariable binds to longer subtrees.
Together, these conditions make $\mathcal{R}$ a non-duplicating, size-decreasing TRS. Termination follows from the length order $\lvert\cdot\rvert$ being a reduction order.
Discovery to $L_{\max} = 7$ has worst-case complexity $O(\alpha^{L_{\mathrm{pattern}} + L_{\mathrm{rep}}})$ in the pattern and replacement lengths and takes $\sim100\,$h on 32 threads, but is a one-time cost: the pre-computed rule set has been published and is efficiently reusable for every task.}

\textbf{Phase 2: Online Pattern Matching \& Cancellation.}
At runtime, rules are loaded and sorted into operator-specific buckets keyed by pattern length and root node.
Explicit rules (those without variables; \emph{ground} in TRS terminology, $\mathcal{R}_g$) are hash-indexed for $O(1)$ lookup while pattern rules (\emph{non-ground}, $\mathcal{R}_v$) are stored as trees for subtree matching.
\camready{Given an input expression $\bm{\tau}$, \textsc{SimpliPy} alternates pattern application and term cancellation up to $K$ times (Algorithm~\ref{alg:simplipy_online}; full version in Appendix~\ref{app:algorithms}).}
\camready{
\begin{algorithm}[!h]
\caption{\textsc{SimpliPy} Simplification (abstract)}
\label{alg:simplipy_online}
\begin{algorithmic}[1]
\STATE \textbf{Input:} expression $\bm{\tau}$, rule set $\mathcal{R}$, max iterations $K = 5$, optional max pattern length $L_\text{max}$
\STATE $\bm{\tau}_{\mathrm{orig}} \gets \bm{\tau}$
\FOR{$i = 1$ to $K$}
    \STATE $\bm{\tau} \gets \textsc{CancelTerms}(\bm{\tau})$ 
    \STATE $\bm{\tau} \gets \textsc{ApplyRules}(\bm{\tau}; \mathcal{R}, L_\text{max})$ 
    \IF{$\bm{\tau}$ unchanged}
        \STATE \textbf{break}
    \ENDIF
\ENDFOR
\STATE Sort commutative operands to canonical order; replace cancellation-introduced literals with the placeholder for free constants $\diamond$.
\STATE \textbf{return} $\bm{\tau}$ if $|\bm{\tau}| \le |\bm{\tau}_{\mathrm{orig}}|$ else $\bm{\tau}_{\mathrm{orig}}$
\end{algorithmic}
\end{algorithm}
}\\
\textsc{ApplyRules} matches recursively from largest patterns to smallest while \textsc{CancelTerms} merges associative-commutative clusters (e.g., $[\text{\token{+}}, \text{\token{x1}}, \text{\token{x1}}] \to [\text{\token{mult2}}, \text{\token{x1}}]$) and cancels inverse pairs.
An optional $L_\text{max}$ caps the longest pattern considered, providing control over the speed/quality trade-off.

\camready{For an input with $n$ nodes, $R$ rules of maximum pattern length $P$, and $K$ iterations, the worst-case online complexity is $O(n^2 \cdot (K \cdot R \cdot P + \log n))$ (derivation in Appendix~\ref{app:algorithms}).}


As an example, \textsc{SimpliPy} simplifies
\begin{equation}
  \left| \frac{x_1}{2} \right|^2 + c_1 \cdot e^{x_2 - x_2} + c_2 \enspace \longrightarrow \enspace c_1 + \left( \frac{x_1}{2} \right)^2
\end{equation}
effectively collapsing the exponential to 1, removing the redundant absolute operator, and substituting $c_1 \gets c_1 + c_2$.

\subsection{Architecture}
\label{sec:architecture}

\textsc{Flash-ANSR} builds on an encoder-decoder Transformer architecture inspired by \citet{biggio2021neural}.
\camready{Fully exploiting \textsc{SimpliPy}'s data-quality and -scale gains requires a base architecture that scales well to large models and training sets.
We therefore depart from \citet{biggio2021neural} and propose an improved, scalable architecture (Table~\ref{tab:vs_biggio}).
}

\camready{
\begin{table}[ht]
  \centering
  \small
  \caption{Architectural and inference choices in \textsc{Flash-ANSR} relative to NeSymReS \cite{biggio2021neural}.}
  \label{tab:vs_biggio}
  \begin{tabular}{@{}lll@{}}
    \toprule
    Component & NeSymReS & \textsc{Flash-ANSR} \\
    \midrule
    Decoder norm        & Post-norm        & Pre-norm \\
    Encoder norm        & LayerNorm / None & m. RMSSetNorm \\
    Positional enc.     & Sinusoidal       & RoPE \\
    Input encoding      & 16-bit IEEE-754  & 32-bit IEEE-754 \\
    Decoding strategy   & Beam Search      & Softmax Sampling \\
    Inference optimizer & BFGS             & Lev.-Marq. \\
    \bottomrule
  \end{tabular}
\end{table}
}

We opt for the pre-RMSNorm Transformer decoder architecture \cite{vaswani2023attentionneed, Zhang2019RootMS, Xiong2020OnLN} with multi-head FlashAttention \cite{dao2023flashattention2fasterattentionbetter}\camready{. Pre-norm stabilizes gradient flow at deeper decoders and is the de-facto choice for modern Transformer scaling}.
To encode data sets of variable size, we use a variant of the Set Transformer \cite{lee2019settransformerframeworkattentionbased} to efficiently encode the input data and use $I = S = 128$ induction and seed points respectively.
Inspired by \citet{Child2019GeneratingLS, Xiong2020OnLN}, and \citet{Zhang2019RootMS, pmlr-v162-zhang22ac}, we introduce the Pre-Norm paradigm to the Set Transformer architecture.
We introduce a masked RMSSetNorm based on \citet{Zhang2019RootMS, pmlr-v162-zhang22ac} to replace the standard LayerNorm \cite{ba2016layernormalization} or SetNorm \cite{pmlr-v162-zhang22ac}, allowing us to standardize over the same number of axes as SetNorm while only requiring half the number of statistics and re-scaling parameters,
and accounting for padding (see Appendix \ref{app:masked_rmsnorm} for derivation).
\camready{In the decoder we replace sinusoidal positional encodings with RoPE \cite{su2023roformerenhancedtransformerrotary}, which injects relative position directly into attention and eliminates the need to recover relative offsets from additive absolute encodings.}

As in NeSymReS \cite{biggio2021neural}, we preprocess inputs to the Set Transformer by representing each scalar by a multi-hot encoding over a fixed bit representation following the IEEE-754 standard \cite{4610935}.
We use 32 bits instead of just 16 bits to represent each scalar value, leading to an input tensor $\mathbf{Z}^{(0)} \in \mathbb{R}^{N \times M \times 32 \cdot (D + 1)}$ with batch size $N$\camready{, broadening the representable range to match physical-domain data ($10^{-38}$ to $10^{38}$ vs $10^{-4}$ to $10^4$ at 16-bit)}.

\subsection{Data Generation}
\label{sec:data_generation}

\camready{Training data is generated fully on-the-fly through a four-stage pipeline: \emph{(1)} skeleton sampling, \emph{(2)} simplification, \emph{(3)} decontamination against the test set, and \emph{(4)} dataset rendering. We illustrate each stage on a single draw and refer to Appendix \ref{app:data_gen_example} for all sampling priors and hyperparameters.}

  \camready{
  \textbf{(1) Skeleton sampling.}
We sample the operator count $n_\text{ops} \in \{0, \dots, 17\}$ from a length-exponential prior $P(n_\text{ops} = k) \propto \exp(k^{0.7})$, build a prefix skeleton $\bm{\bar{\tau}}$ (constants masked with placeholders $\diamond$) with the Lample \& Charton algorithm \cite{lample2019deeplearningsymbolicmathematics}, sample and assign operators to internal nodes (with $\{$\token{+}, \token{-}, \token{*}, \token{/}$\}$ weighted 10$\times$ over other operators), and populate the leaves with a multiset of variables and the constant placeholder \token{$\diamond$}, e.g. with $n_\text{ops} = 4$:
\begin{equation*}
  \bm{\bar{\tau}}^{(1)} = [\text{\token{+}},\, \text{\token{mult2}},\, \text{\token{sin}},\, \token{$x_1$},\, \text{\token{pow2}},\, \token{$\diamond$}].
\end{equation*}
}
\camready{
\textbf{(2) Simplification.}
The candidate is normalized with \textsc{SimpliPy} ($L_\text{max} = 4$) and rejected if simplification produces non-finite symbols. In our example, \textsc{SimpliPy} absorbs \token{pow2} into the constant placeholder, yielding
\begin{equation*}
  \bm{\bar{\tau}}^{(1)*} = [\text{\token{+}},\, \text{\token{mult2}},\, \text{\token{sin}},\, \token{$x_1$},\, \token{$\diamond$}].
\end{equation*}
}
\camready{
\textbf{(3) Decontamination.}
We structurally prune all constant nodes from both the candidate and every held-out test skeleton. We then compare the pruned forms symbolically (token equality) and numerically: each pruned form is evaluated on a fixed grid $X_\text{check} \sim \mathcal{U}(-10, 10)^{512 \times D}$, rounded element-wise to four decimals, and hashed; the candidate is rejected on hash collision with any held-out output. NaN entries are replaced with $0$ before hashing, biasing the filter toward rejection at NaN or near-zero positions. For our example, the pruned form $\bm{\tilde{\tau}}^{(1)} = [\text{\token{mult2}},\, \text{\token{sin}},\, \token{$x_1$}]$ has no test-set match and is accepted for dataset rendering.
}

\camready{
\textbf{(4) Dataset rendering.}
We sample the number of data points $M \sim \mathcal{U}(1, 1024)$, the per-dimension support range from $\mathcal{N}(0, 10)$, and each constant from $\mathcal{N}(0, 5)$, then compile and evaluate $y = f_{\bm{\bar{\tau}}}(X;\bm{c})$. Datasets containing complex, non-finite, or non-numeric values trigger up to four resampling attempts for $\bm{\bar{\tau}}$. Persistent failures reject the entire instance. For our example $\bm{\bar{\tau}}^{(1)*}$, $c_1 = 1.15$, $M = 137$, and the resulting $(X, y)$ is finite-valued and accepted.
}
\subsection{Training}
We minimize the cross-entropy loss over the joint distribution of normalized skeletons and datasets $p(\bm{\bar{\tau}}^*, \mathcal{D})$:
\begin{equation}
  \hat \theta = \arg \min_\theta \mathbb{E}_{\substack{\bm{\bar{\tau}} \sim p(\bm{\bar{\tau}} )\\ \mathcal{D} \sim p(\mathcal{D} \given \bm{\bar{\tau}})}} \left [ - \sum_{t=1}^{L} \log p_{\theta} \left (\bar{\tau}_t^* \given \bar{\tau}_{<t}^*, \mathcal{D} \right ) \right ]
\end{equation}
where $p_{\theta}$ is the likelihood assigned by the Transformer decoder at step $t$, conditioned on the previous tokens $\bar{\tau}_{<t}^*$ and the dataset $\mathcal{D}$ encoded by the Set Transformer.
Encoder and decoder are trained jointly and end-to-end.
We train four models (3M, 20M, 120M, and 1B parameters) using the hyperparameters listed in Appendix \ref{app:hyperparameters}.

\subsection{Inference}
\label{sec:inference}

\camready{
The posterior $p(\bm{\bar{\tau}} \given \mathcal{D})$ is multi-modal, meaning that multiple distinct functional hypotheses and forms can fit the same data.
For diverse candidate generation and better exploration, we replace beam search with softmax sampling and draw $K$ \emph{independent} candidate skeletons $\bm{\bar{\tau}}_k \sim p_{\hat{\theta}}(\bm{\bar{\tau}} \given \mathcal{D})$.
}
Integer multiplication and division operator tokens in $\bm{\bar{\tau}}$ are replaced with multiplications by free constants at a slightly increased optimization cost to allow for fine-grained numerical optimization of the expression, e.g. $[..., \text{\token{mult2}}, \text{\token{$\cdot$}}, ...] \to [..., \text{\token{*}}, \text{\token{$\diamond$}}, \text{\token{$\cdot$}}, ...]$.
All generated skeletons are then simplified with SimpliPy ($L_\text{max} = 4$), deduplicated symbolically, and compiled before their constants are optimized with the Levenberg--Marquardt algorithm \cite{Levenberg1944AMF, marquardt1963} under a least-squares objective using SciPy.optimize.curve\_fit \cite{SciPy-NMeth2020}, yielding optimized expressions $\bm{\hat{\tau}}$.
\camready{Empirically, LM achieves ${\sim}10$\,pp higher vNRR than BFGS \cite{broyden1970, fletcher1970, goldfarb1970, shanno1970} on \textsc{FastSRB} at matched compute (Appendix~\ref{app:refiner_choice}).}
Optimizations are run in parallel on multiple CPU cores and are restarted 8 times with initial values sampled from $c_0 \sim \mathcal{N}(0, \sigma=5)$, matching the training prior.
Inspired by many other methods \cite{biggio2021neural, cranmer2023interpretablemachinelearningscience}, the predicted expressions are then sorted by their fit quality and a parsimony regularization
\begin{equation}
\label{eq:parsimony_scoring}
\bm{\hat{\tau}}^\star = \arg \min_{\bm{\hat{\tau}}} \, \log_{10} \text{FVU}(\bm{\hat{\tau}}) + \gamma \cdot |\bm{\hat{\tau}}|
\end{equation}
where $\text{FVU}(\bm{\hat{\tau}})$ is the fraction of variance unexplained (Equation \ref{eq:fvu}) achieved by an expression $\bm{\hat{\tau}}$, and $\gamma$ is a small constant penalizing the complexity of a predicted expression. 
Thus, an expression $\bm{\hat{\tau}}_b$ is considered better than an expression $\bm{\hat{\tau}}_a$ if
\begin{equation}
\frac{\text{FVU}(\bm{\hat{\tau}}_b)}{\text{FVU}(\bm{\hat{\tau}}_a)} < 10^{-\gamma \left(|\bm{\hat{\tau}}_b| - |\bm{\hat{\tau}}_a|\right)} \text{.}
\end{equation}
By default, we set $\gamma = 0.05$, meaning that each additional symbol should decrease the FVU by $11\%$ relative to the expression without that symbol.

\section{Experiments}

We evaluate of \textsc{Flash-ANSR} against established baselines in amortized SR (NeSymReS, E2E) and direct optimization (PySR) on 115 expressions and data derived from the \textsc{FastSRB} benchmark \citep{martinek2025fastsymbolicregressionbenchmarking}\footnote{\camready{We exclude 5 equations (B4, B7, B17, II.24.17, III.14.14) whose prescribed sampling regimes induce \texttt{sqrt}-domain or \texttt{exp}-overflow issues (see Appendix~\ref{app:excluded_equations} for details).}}.
Compared to other commonly used benchmarks (e.g. Feynman \citep{udrescu2020aifeynmanphysicsinspiredmethod}, Nguyen \citep{hoai2002solving, johnson2009genetic, keijzer2003improving, uy2011semantically, petersen2021deep}), which restrict the domain of input data to small and arbitrary intervals around zero, \textsc{FastSRB} adopts the physically motivated data domains of SRSD \cite{matsubara2024rethinkingsymbolicregressiondatasets}, ensuring that methods are evaluated on data distributions that reflect real scientific applications.

We evaluate under strict machine-precision recovery metrics and analyze performance as a function of three key scaling axes: test-time compute, data sparsity, and noise robustness.

\subsection{Time-Normalized Recovery Rate}
\label{sec:exp_test_time_compute}

We take inspiration from \citet{biggio2021neural} and evaluate the inference-time-recovery-rate Pareto frontier by sweeping over a characteristic scaling parameter for each method (Table \ref{tab:test_time_compute_sweep}), varying the inference time budget from the millisecond regime ($< 0.1$s) up to a medium-horizon search ($\approx 1000$s) in powers of 2.
No artificial restrictions are imposed on hardware acceleration (see Appendix \ref{app:test_time_compute_system}), and no model is stopped early during inference.
For each test expression, we evaluate on 10 distinctly sampled datasets for short-term runs and 5 for medium-horizon runs.
Baselines are configured as close to their default settings as possible (see Appendix \ref{app:baselines} for details).

\begin{table}[ht]
  \caption{
    Ranges for characteristic inference-time scaling parameters used to analyze performance across orders of magnitude in inference time.
    Evaluation of our 1B parameter model (Appendix \ref{app:metrics}) is limited to 64k samples to account for additional inference time induced by the larger model size.
    }
  \label{tab:test_time_compute_sweep}
  \vskip 0.1in
  \centering
  \begin{small}
  \begin{tabular}{@{}lllr@{}}
    \toprule
    \textsc{Method} & \textsc{Parameter} & \multicolumn{2}{c}{\textsc{Ranges}} \\
    \cmidrule(l){3-4}
           &           & Short ($10 \times$) & Medium ($5 \times$) \\
    \midrule
    NeSymReS   & Beam Width  & $1 \dots 32$   & $128, 512$ \\
    E2E        & Beam Width  & $1 \dots 512$ & -- \\
    PySR       & Iterations  & $1 \dots 1024$ & $4096, 16\text{k}$ \\
    \midrule
    Flash-ANSR & Samples & $1 \dots 16\text{k}$ & $64\text{k}, 256\text{k}$ \\
    \bottomrule
  \end{tabular}
  \end{small}
\end{table}

\subsection{Data Sparsity}
\label{sec:exp_data_sparsity}

We evaluate the ability of \textsc{Flash-ANSR} to recover physical laws from sparse data by varying the number of points to fit, $M \in \{2^0, \dots, 2^{11}\}$.
To ensure a fair comparison, we fix the inference budget to approximately 10 seconds per expression for all methods.
The specific hyperparameters corresponding to this budget are detailed in Table \ref{tab:fixed_test_time_compute_settings}.

\begin{table}[ht]
  \caption{Fixed test time compute settings for the sparse data and noisy data experiments.}
  \label{tab:fixed_test_time_compute_settings}
  \vskip 0.1in
  \centering
  \begin{small}
  \begin{tabular}{@{}llr}
    \toprule
    \textsc{Method} & \textsc{Parameter} & \textsc{Value} \\
    \midrule
    NeSymReS   & Beam Width  & 4 \\
    E2E        & Beam Width  & 256 \\
    PySR       & Iterations  & 128 \\
    \midrule
    Flash-ANSR v23.0-3M & Samples & 4096 \\
    Flash-ANSR v23.0-20M & Samples & 2048 \\
    Flash-ANSR v23.0-120M & Samples & 2048 \\
    Flash-ANSR v23.0-1B & Samples & 512 \\
    \bottomrule
  \end{tabular}
  \end{small}
\end{table}

\subsection{Noisy Data}
\label{sec:exp_noisy_data}

To assess robustness to noise, we introduce additive Gaussian noise to the target variables following a protocol similar to \citet{lacava2021contemporarysymbolicregressionmethods}.
For the relative noise levels $\eta \in \{10^{-3}, 10^{-2}, 10^{-1}\}$, we corrupt the data via
\begin{equation}
    \tilde{y}_m = y_m + \epsilon_m, \quad \epsilon_m \sim \mathcal{N}(0, \sigma=(\eta \cdot \sigma_y))
\end{equation}
where $\sigma_y$ is the standard deviation of the clean targets.

\subsection{Metrics}
\label{sec:metrics}

We assess model performance based on two central metrics capturing the goodness of fit and the parsimony of the predicted expressions.

Relying on exact symbolic match criteria for evaluation is fundamentally flawed for two reasons.
Incomplete symbolic canonicalizations of functionally equivalent expressions are not identified as such without expensive manual intervention, leading to misleadingly low recovery rates.
Purely symbolic criteria also disregard the vast amount of diverse, functionally distinct expressions that can fit the data equally well as the ground truth.
Hence, we focus on numeric recovery as the primary criterion for success.

\textbf{Numeric Recovery Rate (NRR).}
Since problems can have widely varying output scales, we base our numeric criterion on the Fraction of Variance Unexplained (FVU) between the ground-truth values $y$ and the values obtained from a predicted expression $\hat{y}$.
\camready{NRR is the fraction of test \emph{problems} for which the model's best candidate (Eq.~\ref{eq:parsimony_scoring} for our models) achieves $\mathrm{FVU}(y, \hat{y}) \le \epsilon_{32} \approx 1.19 \times 10^{-7}$ (32-bit machine precision).}
We further distinguish between the recovery rate on the fit data split (fNRR) and an unseen validation data split of equal size (vNRR) for each test instance.

\textbf{Expression Length Ratio.}
As a proxy for symbolic quality and interpretability, we evaluate the parsimony of the predicted expressions using the ratio of lengths between predicted and ground truth expressions $|\bm{\hat{\tau}}| / |\bm{\tau}|$, where $|\bm{\tau}|$ denotes the number of nodes in the expression tree of $\bm{\tau}$.

\camready{

\textbf{Total Nestedness.} We complement the above with a structural measure, counting the total operator-chaining depth.

\textbf{Rewrite Fraction.} To estimate diversity, we compute the share of generated candidates $\bm{\tau}_k$ per problem that are rewrite-equivalent under \textsc{SimpliPy}-normalization.

\textbf{Bloat.} We report $\mathbb{E}_{\bm{\tau}_k}[|\bm{\tau}_k|/|\textsc{SimpliPy}(\bm{\tau}_k)|] - 1$, measuring the inflation of candidates relative to their simplified forms.
}

We indicate the objective of each metric, e.g. $\text{NRR}\uparrow^{[0, 100]}$ and use bootstrapping to estimate the 95\% confidence intervals for all statistics \cite{riezler2024validity}.

\subsection{Symbolic Simplification Efficiency}
We directly compare the effectiveness and efficiency of simplifications with SymPy \cite{sympy} and \textsc{SimpliPy} based on $2^{16}$ (64k) expressions sampled from our training distribution (Section \ref{sec:data_generation}).
For SymPy, we implement a compatibility layer that converts our tokenized prefix representation to the corresponding infix string.
Then, to work around SymPy's inherent inability to combine symbols representing free constants, we temporarily substitute randomly sampled values $c_k \sim \mathcal{U}(-10, 10)$ into the expression before parsing it with SymPy, measuring the time of its \texttt{simplify} routine, and masking numerical values again with the placeholder $\diamond$.
The simplified expression is then converted back to our prefix format with \textsc{SimpliPy}'s parser.

\section{Results}
\label{sec:results}

\subsection{Time-Normalized Recovery Rate}
\label{sec:results_test_time_compute}

\begin{figure}[ht]
  \vskip -0.1in
  \begin{center}
    \centerline{\includegraphics[width=\columnwidth]{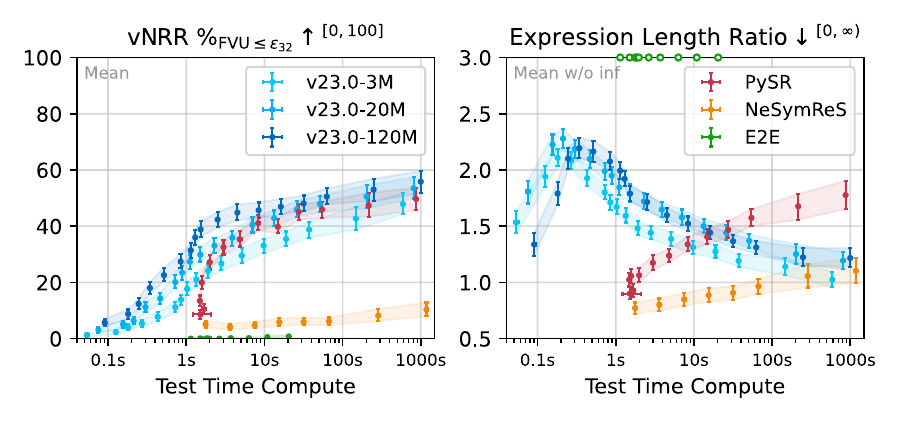}}
    \caption{
        \textbf{Left:} Validation Numeric Recovery Rate (vNRR) as a function of inference time (log scale). \textsc{Flash-ANSR} models (shades of blue) scale monotonically with compute, with the 120M model partially surpassing the PySR baseline (red). Baselines NeSymReS and E2E fail to generalize to the benchmark.
        \textbf{Right:} Expression Length Ratio $|\bm{\hat{\tau}}|/|\bm{\tau}|$ versus compute. We observe a \textit{parsimony inversion}: while PySR increases complexity to minimize error over time, \textsc{Flash-ANSR} converges toward simpler, more canonical expressions as the sampling budget increases. Shaded regions denote 95\% confidence intervals.
    }
    \label{fig:small_test_time_compute_fastsrb}
    \vspace{-15pt}
  \end{center}
\end{figure}

We present the results of the test-time normalized performance on the \textsc{FastSRB} benchmark in Figure \ref{fig:small_test_time_compute_fastsrb}.

\textbf{Baselines.}
Consistent with the limitations identified in Section \ref{sec:related_work}, prior amortized methods struggle to generalize to the realistic domains of \textsc{FastSRB}.
NeSymReS saturates at a recovery rate of approximately $10\%$, while E2E fails to achieve meaningful recovery ($< 2.5\%$) across all budgets.

\textbf{Accuracy Scaling.}
\textsc{Flash-ANSR} demonstrates strong power-law scaling with respect to inference compute.
While the smallest model (3M) lags behind the genetic programming baseline, the larger variants and the 1B model (Appendix \ref{app:metrics}) effectively bridge the gap.
Notably, the 120M parameter model achieves parity with PySR in the medium-compute regime ($\approx 10$s) and partially surpasses it at higher budgets, reaching a peak recovery rate of $\sim58\%$ compared to PySR's $50.0\%$.
This shows that with sufficient scale and sampling budget, an amortized prior can match or exceed the search efficiency of mature evolutionary algorithms.

\textbf{The Parsimony Inversion.}
Crucially, we observe a fundamental divergence in how the methods navigate the accuracy-complexity trade-off (Figure \ref{fig:small_test_time_compute_fastsrb}, right).
PySR employs a sophisticated model selection strategy: it maintains a Pareto front of the best expressions found at every complexity level and selects the final candidate by maximizing the negative gradient of the log-loss with respect to complexity (finding the ``knee'' of the curve).
Without any parsimony penalty beyond this default strategy, however, PySR exhibits a positive correlation between time and expression complexity: as the evolutionary search progresses, the length ratio increases from $0.94$ to $1.85$.
This suggests that even with parsimony pressure, the GP optimizes residual error by appending corrective terms, resulting in expressions that fit the data via complexity rather than structural discovery.

Conversely, \textsc{Flash-ANSR} exhibits an \textit{inverse} scaling trend.
As the sampling budget increases, the average length ratio of the best-found solution \textit{decreases}, converging towards the ground truth length (e.g., the 120M model improves from $1.40$ to $1.27$), while simultaneously improving recovery rates.
With more compute, the model successfully samples these rarer, concise ``needles in the haystack'', rather than constructing complex approximations.

\begin{figure}[ht]
  \vskip -0.1in
  \begin{center}
    \centerline{\includegraphics[width=\columnwidth]{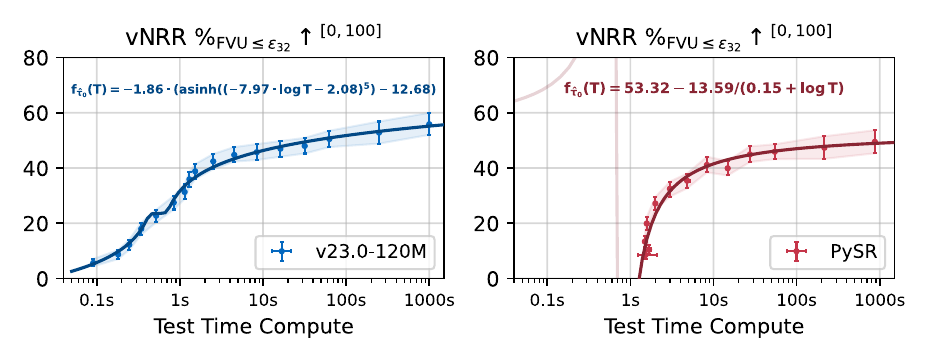}}
    \caption{
      \textsc{Flash-ANSR} fits to its own and PySR's scaling curves $\log_{10}(T)$ vs.\ vNRR from Figure \ref{fig:small_test_time_compute_fastsrb} (left) using v23.0-120M, $\gamma = 0.15$, 128k choices $\approx$ 10 min.
      Extrapolation suggests an asymptotic vNRR $\propto \log \log T$ scaling for \textsc{Flash-ANSR}, and an asymptotic upper limit for PySR around $53\%$.
    }
    \label{fig:fitted_curves_test_time_compute_fastsrb}
    \vspace{-25pt}
  \end{center}
\end{figure}

As a realistic application of our method, we present the identification of mathematical test-time scaling laws on the \textsc{FastSRB} benchmark in Figure \ref{fig:fitted_curves_test_time_compute_fastsrb}.

\subsection{Symbolic Simplification Efficiency}
\label{sec:symbolic_simplification_efficiency}

\begin{figure}[ht]
  \vskip -0.1in
  \begin{center}
    \centerline{\includegraphics[width=\columnwidth]{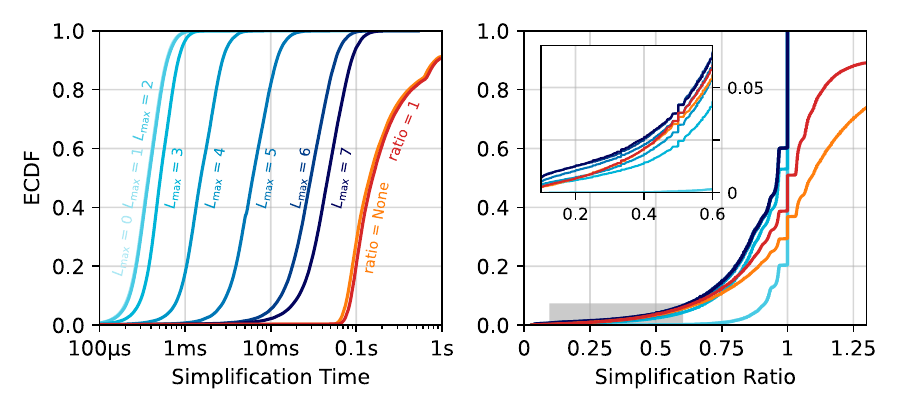}}
    \caption{
      \textbf{Left:} Empirical Cumulative Distribution Functions (ECDFs) of simplification wall-clock time. Our \textsc{SimpliPy} rewriting engine (shades of blue, varying $L_{\max}$) operates in the low to moderate millisecond regime, orders of magnitude faster than the SymPy baseline (orange, red).
      \textbf{Right:} ECDF of the simplification ratio $|\bm{\tau^*}| / |\bm{\tau}|$. The inset highlights the tail of the distribution. Our method with $L_{\max} \ge 5$ achieves simplification ratios comparable to the SymPy baseline while maintaining high throughput.
    }
    \label{fig:simplification_comparison_simplipy_sympy}
  \vspace{-20pt}
  \end{center}
\end{figure}

Figure \ref{fig:simplification_comparison_simplipy_sympy} characterizes the performance of our \textsc{SimpliPy} engine and the SymPy \texttt{simplify} routine.

\textbf{Speedup.}
Our measurements (Figure \ref{fig:simplification_comparison_simplipy_sympy}, left) reveal a significant acceleration with respect to the SymPy baseline which exhibits a median simplification time of approximately $100$ms per expression, rendering it prohibitive for real-time applications or large-batch training.
In contrast, our pattern-based engine operates in the millisecond regime for $L_{\max}=4$ (used to train and evaluate \textsc{Flash-ANSR}).

\textbf{Quality Preservation.}
Critically, this speed only comes at a minimal cost of simplification quality.
Figure \ref{fig:simplification_comparison_simplipy_sympy} (right) plots the distribution of the simplification ratio $|\bm{\tau}^*| / |\bm{\tau}|$.
We observe that the distributions for $L_{\max} \ge 5$ even exceed the baseline in terms of conciseness, indicating that our local rewriting rules and cancellations capture the vast majority of reductions found by the more complex heuristics of SymPy.
While \textsc{SimpliPy} results in strictly shorter or at most equal lengths, SymPy's simplification paradoxically increases the length of about $38\%$ to $52\%$ of expressions (even with the ``ratio'' parameter set to 1), and fails to complete within a 1-second timeout for $9\%$ of expressions.

\subsection{Data Sparsity}
\label{sec:results_data_sparsity}

\begin{figure}[ht]
  \vskip -0.1in
  \begin{center}
    \centerline{\includegraphics[width=\columnwidth]{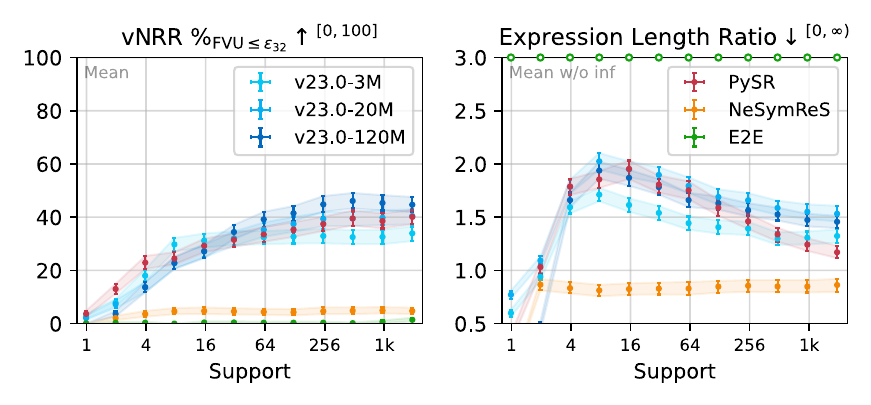}}
    \caption{
        \textbf{Left:} Validation Numeric Recovery Rate (vNRR) vs. the number of support points $M$. \textsc{Flash-ANSR} (120M) outperforms PySR in the dense data regime ($M > 64$).
        \textbf{Right:} Expression Length Ratio $|\bm{\hat{\tau}}|/|\bm{\tau}|$. We observe a distinct ``Complexity Peak'' at $M \approx 8$, where the model generates expressions significantly longer than the ground truth.
        This peak coincides with a regime of high uncertainty (low log-probability) and excess constant usage, suggesting the model is \textit{interpolating} the sparse points via complex aliasing rather than identifying the underlying law.
    }
    \label{fig:small_n_support_fastsrb}
    \vspace{-15pt}
  \end{center}
\end{figure}

We investigate the impact of data sparsity on model performance by varying the number of support points $M$ from $1$ to $2048$ (Figure \ref{fig:small_n_support_fastsrb}).
While the amortized baselines (NeSymReS, E2E) remain largely ineffective across all regimes, \textsc{Flash-ANSR} and PySR demonstrate strong scaling behavior.

The structural evolution of the predicted expressions reveals a phase transition analogous to \emph{(Deep) Double Descent} \cite{Belkin2018ReconcilingMM, nakkiran2019deepdoubledescentbigger}.
At the critical threshold of $M \approx 8$ points, we observe a ``complexity peak'' where the average expression length overshoots the ground truth by up to 2 times.
This correlates with a surge in excess constants (using degrees of freedom to minimize the fit error) and a dip in the model's log-probability (Appendix \ref{app:data_sparsity}),
suggesting a transition through three distinct regimes:

\textbf{The Parsimonious Regime ($M < 8$):}
The data is insufficient to contradict the model's bias toward simplicity and the scoring function's preference for parsimonious expressions (Equation \ref{eq:parsimony_scoring}),
resulting in confident, concise approximations with high bias but low variance.

\textbf{The Interpolation Regime ($M \approx 16$):}
The data density is sufficient to rule out trivial forms but insufficient to constrain the search to the true symbolic law.
To maintain low fit error, the method saturates its complexity budget, inserting excess constants and operators to construct ``aliasing'' approximations.
The drop in log-probability confirms this as a regime of high uncertainty, where probability mass is spread across many complex candidates.

\textbf{The Identification Regime ($M \to 2048$):}
The symbolic cost of explaining the data via complex approximation exceeds the model's generative capabilities, encouraging a posterior collapse to the ground truth -- the only remaining solution that fits the data and is concise enough to be predicted by the model.
We observe that distribution shifts in the data can disrupt this delicate balance, preventing successful identification even in the dense data regime.

\subsection{Noisy Data}

\begin{figure}[ht]
  \vskip -0.1in
  \begin{center}
    \centerline{\includegraphics[width=\columnwidth]{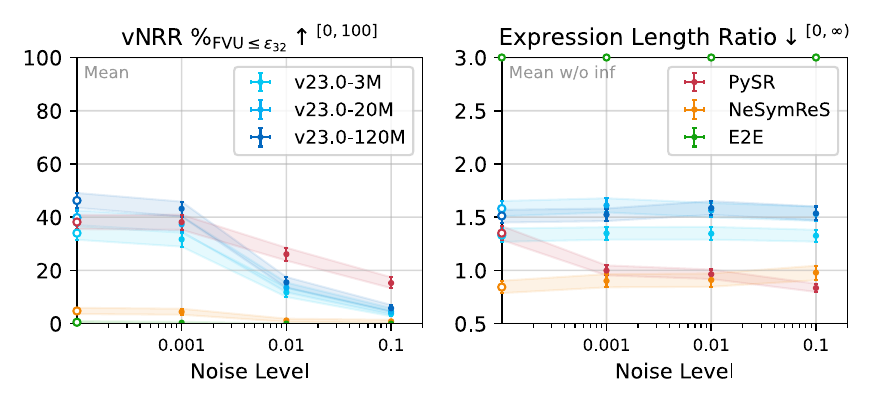}}
    \caption{
        \textbf{Left:} Validation Numeric Recovery Rate (vNRR) vs. the noise level $\eta$ (log scale). While \textsc{Flash-ANSR} (shades of blue) achieves competitive performance in the noiseless regime (circles), PySR (red) retains higher recovery rates at larger noise levels.
        \textbf{Right:} Expression Length Ratio $|\bm{\hat{\tau}}|/|\bm{\tau}|$. On noisy data, PySR (red) and NeSymReS (orange) produce expressions of similar complexity to the ground truth whereas \textsc{Flash-ANSR} consistently predicts longer expressions without any apparent trend.
    }
    \label{fig:small_noise_scaling_fastsrb}
    \vspace{-20pt}
  \end{center}
\end{figure}

We demonstrate the effects of distribution shifts induced by additive Gaussian noise on the target values in Figure \ref{fig:small_noise_scaling_fastsrb}.

In low-noise settings, \textsc{Flash-ANSR} demonstrates superior performance with the 120M parameter model surpassing the genetic programming baseline PySR (Figure \ref{fig:small_noise_scaling_fastsrb}, left).
At higher noise levels $\eta \geq 10^{-2}$, we observe a distinct performance crossover.
Here, PySR demonstrates better robustness, retaining a higher recovery rate than \textsc{Flash-ANSR}.

Since \textsc{Flash-ANSR} was trained exclusively on noiseless data, the injection of noise constitutes a distributional shift.
The encoder, having learned that all variance in $y$ carries semantic meaning, misinterprets the noise as high-frequency signals.
Consequently, it attempts to ``explain'' the noise by continuing to generate expressions with rich functional structures, none of which can perfectly fit the intricate variations induced by the noise (Figure \ref{fig:small_noise_scaling_fastsrb}, right).

\subsection{Ablations}
\label{sec:ablations}

\begin{figure}[ht]
  \vskip -0.05in
  \begin{center}
    \centerline{\includegraphics[width=\columnwidth]{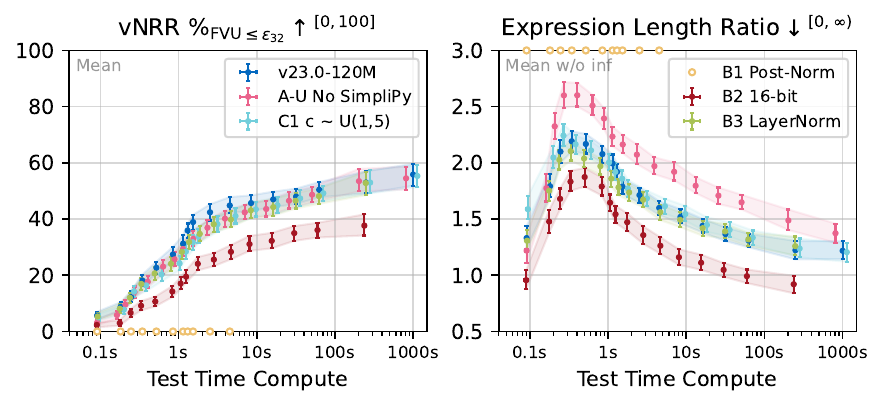}}
    \caption{
      \camready{Test-time compute scaling on \textsc{FastSRB} for ablations A-U, B1, B2, B3, C1 of the 120M model.
      \textbf{Left:} vNRR. Both training-data ablations trail the baseline by a few percentage points up to $\sim$10\,s of inference and converge at higher budgets.
      \textbf{Right:} Expression Length Ratio. The unsimplified ablation produces persistently longer candidates, settling $\sim$40--50\% above canonical at high compute;
      the narrow-prior ablation tracks baseline length.}
    }
    \label{fig:training_data_ablations_120M}
  \vspace{-15pt}
  \end{center}
\end{figure}

\camready{We ablate four axes of our deployed v23.0-120M model:

\textbf{Simplification}: We compare performance using SimpliPy, SymPy \cite{sympy} and no simplification (\textit{A-U}) during training (10M and 100M expressions) and inference.\\
\textbf{Training Prior}: We revert the prior for constants from $\mathcal{N}(0,5)$ to $\,\mathcal{U}(1,5)$ used in \cite{biggio2021neural} (\textit{C1}). \\
\textbf{Architecture}: We measure architectural contributions by reverting individual components to their prior-work baselines \cite{biggio2021neural}: Post-Norm transformers (\textit{B1}), 16-bit input encoding (\textit{B2}), and LayerNorm transformers (\textit{B3}). \\
\textbf{Decoding}: We compare softmax sampling vs.\ beam search on the v23.0-120M baseline on equal sample budgets.

Figure \ref{fig:training_data_ablations_120M} highlights the importance of expression simplification (A-U) and full 32-bit input encoding (B2) to achieve concise predictions and high recovery rates.
}

\camready{
The Post-Norm ablation (B1) training failed due to instability, corroborating our use of the more stable Pre-Norm architecture.
Further training prior (C1) and architecture ablations (B1, B3) have more marginal effects with the largest difference in vNRR between 1s and 10s, beyond which all models saturate at similar performance, suggesting that they struggle with the same hard problems. 
}

\begin{figure}[ht]
  \centering
  \centerline{\includegraphics[width=\columnwidth]{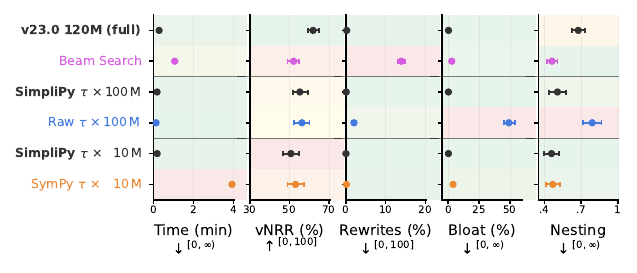}}
  \caption{
    \camready{Hypothesis-space fingerprint at fixed training (full, 100M, 10M) and inference volume ($c = 4096$) on \textsc{FastSRB}.
    Beam Search yields $70\times$ more equivalent rewrites than softmax sampling and recovers 9.4\,pp fewer problems at 4$\times$ wall time.
    Omitting simplification matches \textsc{SimpliPy} on vNRR but produces longer expressions (\emph{Bloat}), and $21\times$ more rewrites.
    \textsc{SymPy} achieves similar performance as \textsc{SimpliPy} at $20\times$ slower inference.
  }}
  \label{fig:fingerprint_main}
  \vspace{-10pt}
\end{figure}

\camready{

Figure~\ref{fig:fingerprint_main} indicates higher diversity and recovery rates in less time using softmax sampling compared to beam search. This is due to rewrite-mode collapse: at $c{=}4096$, beam search retains only 33\% of its nominal budget as distinct canonical hypotheses (48\% for softmax), spending the remainder on syntactic variants of the same dominant canonical form. The deficit concentrates on \emph{short} target equations ($\Delta$vNRR $\approx +17$\,pp); on long ones both decoders explore wide hypothesis sets and the gap shrinks to ${\sim}2$\,pp, suggesting beam search's mode-seeking behavior is most damaging when the posterior is sharply peaked.


}

\section{Conclusion}
\label{sec:conclusion}

We introduced \textsc{SimpliPy}, a high-throughput simplification engine that breaks a key bottleneck preventing amortized SR from scaling to scientific complexity, and \textsc{Flash-ANSR}, which leverages it for large-scale training on diverse, high-quality expressions and efficient inference.
On the \textsc{FastSRB} benchmark, \textsc{Flash-ANSR} dominates existing amortized baselines and matches state-of-the-art genetic programming (PySR) on the inference-time vs.\ recovery-rate Pareto frontier, recovering more concise expressions as the inference budget grows.
Our ablations (Section~\ref{sec:ablations}) further reveal that \textsc{SimpliPy}'s contribution is a \emph{system-component effect} on hypothesis-space coverage: training-data simplification governs surface-form parsimony and reduces compute on canonical-equivalent rewrites.

Limitations remain, particularly robustness to noisy data, which constitutes an out-of-distribution shift for our models.
Future work will incorporate noisy data into training, widen the generated training distributions, and explore alternative encoding and decoding paradigms.
By establishing rigorous evaluation standards and demonstrating scalable performance, our work positions amortized SR as a high-speed complement to classical optimization in scientific discovery.

\section*{Acknowledgments}

This work was funded by the Deutsche Forschungsgemeinschaft (DFG, German Research Foundation) within the Priority Programme SPP 2331 (Machine Learning in Chemical Engineering) -- Project number 466528284.

\section*{Impact Statement}

This paper presents work whose goal is to advance the field of Machine Learning.
There are many potential societal consequences of our work, none of which we feel must be specifically highlighted here.

\bibliography{_paper}
\bibliographystyle{icml2026}

\newpage
\appendix
\onecolumn

\section{Examples}
\label{app:examples}

\begin{figure}[!ht]
  \begin{center}
    \centerline{\includegraphics[width=0.95\columnwidth]{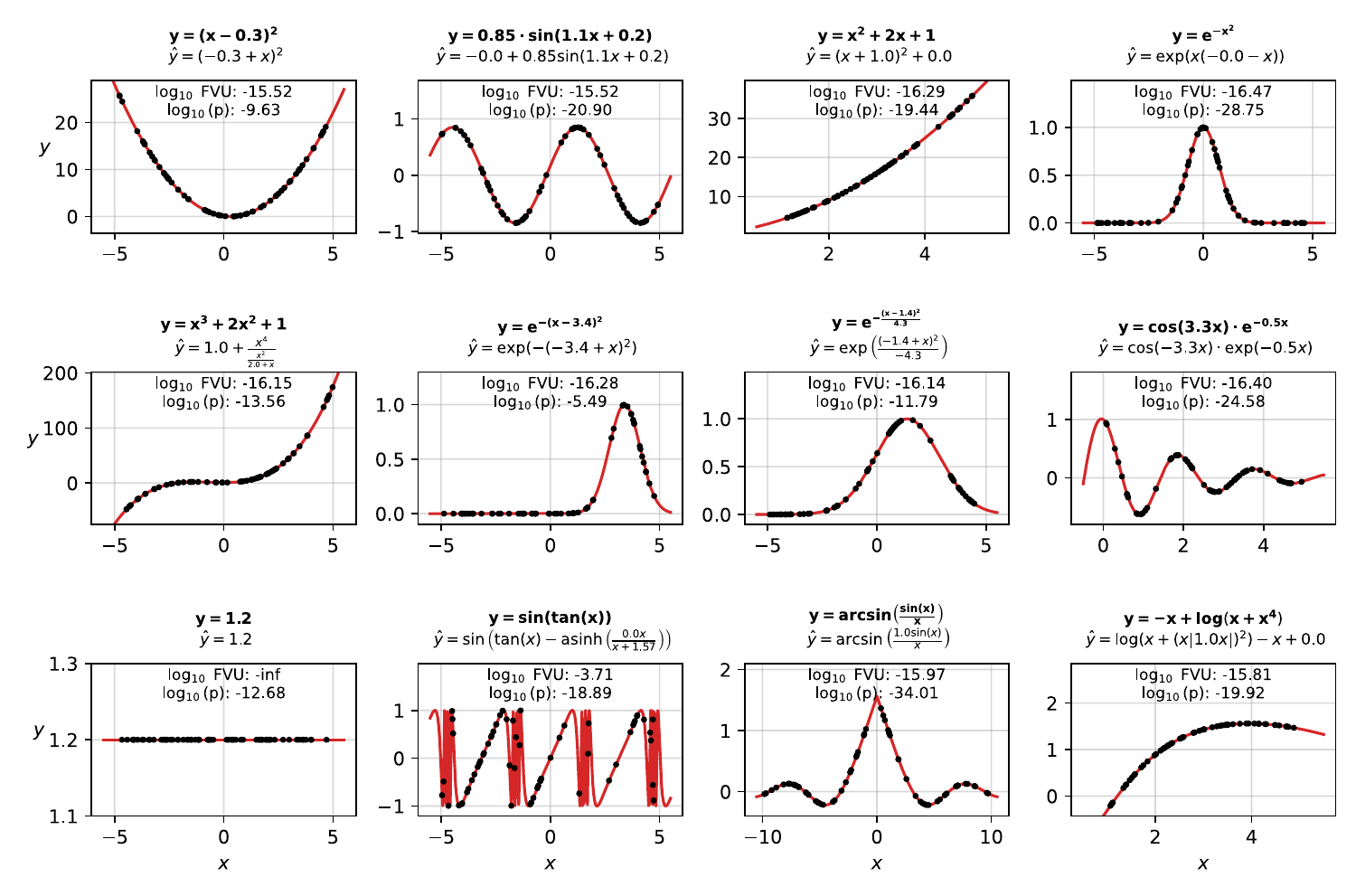}}
    \caption{
        \textsc{Flash-ANSR} fits (v23.0-120M, $\gamma=0.05$, 32k samples $\approx 100$s) (red curves) on uniformly sampled data points (black dots) from several 1D expressions ($M=64$, $\eta = 0$).
        Many ground truth expressions (bold) are recovered exactly, while others are recovered in alternative forms (e.g. $\hat{y} = (x + 1)^2$ for $y = x^2 + 2x + 1$).
    }
    \label{fig:1d_positive_examples}
    \vspace{-10pt}
  \end{center}
\end{figure}

\begin{figure}[!ht]
  \begin{center}
    \centerline{\includegraphics[width=0.5\columnwidth]{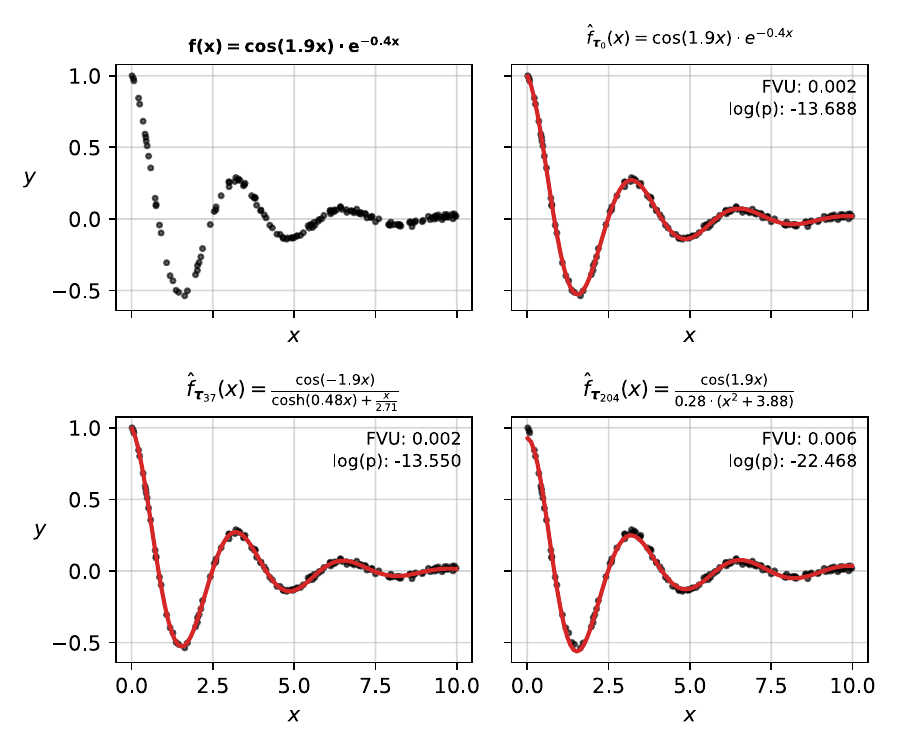}}
    \caption{
      Diverse \textsc{Flash-ANSR} solutions (v23.0-120M, $\gamma=0.05$, 32k samples $\approx 100$s) (red curves) on uniformly sampled data points (black dots) depicting the trajectory of a damped harmonic oscillator ($M=150$, $\eta \approx 0.04$).
      Multiple distinct expressions fit the data equally well given the noisy data. The ground truth expression (top left, bold) is exactly recovered with high probability (top right).
      Other solutions (bottom row) achieve similar fit quality while differing structurally.
    }
    \label{fig:1d_diverse_example_1}
    \vspace{-10pt}
  \end{center}
\end{figure}

\newpage

\begin{figure}[ht]
  \begin{center}
    \begin{subfigure}[t]{0.46\columnwidth}
      \centering
      \includegraphics[width=\linewidth,trim=0 1220 0 10,clip]{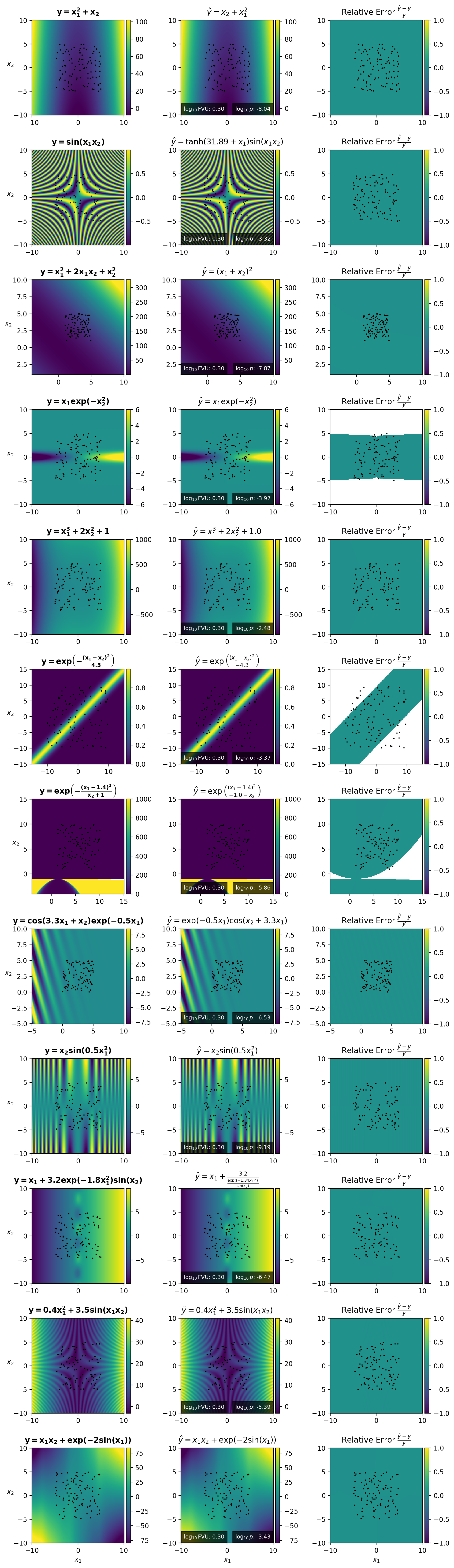}
    \end{subfigure}\hspace{0.07\columnwidth}
    \begin{subfigure}[t]{0.46\columnwidth}
      \centering
      \includegraphics[width=\linewidth,trim=0 5 0 1200,clip]{images/2d_positive_examples.png}
    \end{subfigure}
    \caption{
      \textsc{Flash-ANSR} fits (v23.0-120M, $\gamma=0.05$, 32k samples $\approx 100$s) (middle columns) on uniformly sampled data points (black dots) from several 2D expressions from \citet{biggio2021neural} ($M=100$, $\eta = 0$).
      Many ground truth expressions (left columns, bold) are recovered exactly, while a few others have additional terms that do not affect the relative fit error (right columns).
    }
    \label{fig:2d_positive_examples}
    \vspace{-10pt}
  \end{center}
\end{figure}

\section{Hyperparameters}
\label{app:hyperparameters}

We provide a detailed specification of the hyperparameters and training configurations in Table \ref{tab:hyperparameters}.
\paragraph{Training.}
All models are trained end-to-end using the AdamW optimizer \citep{loshchilov2019decoupledweightdecayregularization} with $\beta_1 = 0.9$.
We utilize a Warmup-Stable-Decay learning rate schedule \citep{hu2024minicpmunveilingpotentialsmall, Hgele2024ScalingLA} with a linear warmup from $0$ to $\eta_{\text{max}}$, a constant phase, and a linear decay to $0$.
The 3M, 20M, and 120M models use a $10\%/70\%/20\%$ split while the 1B model uses an extended $18.75\%/56.25\%/25\%$ split.
We apply gradient clipping with a maximum global norm of $2.0$.


\paragraph{Architecture \& Implementation.}
Our models are implemented in PyTorch 2.9.
To maximize training throughput, we use Automatic Mixed Precision (AMP) with \texttt{torch.compile} \citep{ansel2024pytorch}. The AMP dtype is bf16 on Ada / Ampere / Hopper GPUs (RTX 4090, A100, H200) and fp16 with loss scaling on the Turing RTX 2080Ti used for the 3M model.
We apply dropout with $p=0.1$ to attention weights and to the output of feed-forward networks.
In the decoder, we utilize Rotary Positional Embeddings (RoPE) \citep{su2023roformerenhancedtransformerrotary} applied to the queries and keys of the self-attention mechanism.
The Set Transformer encoder uses $I=128$ inducing points and $S=128$ seed vectors across all model sizes.

\begin{table}[!ht]
  \caption{Hyperparameters for the Flash-ANSR model series.}
  \label{tab:hyperparameters}
  \begin{center}
    \begin{small}
      \begin{tabular}{lcccc}
        \toprule
        \textsc{Hyperparameter} & \textsc{v23.0-3M} & \textsc{v23.0-20M} & \textsc{v23.0-120M} & \textsc{v23.0-1B} \\
        \midrule
        \multicolumn{5}{l}{\textit{Encoder}} \\
        Dimension $d_{enc}$ & 192 & 384 & 640 & 1280 \\
        Heads & 3 & 6 & 10 & 20 \\
        ISAB Layers & 1 & 2 & 5 & 10 \\
        SAB Layers & 1 & 2 & 5 & 10 \\
        \midrule
        \multicolumn{5}{l}{\textit{Decoder}} \\
        Dimension $d_{dec}$ & 192 & 384 & 640 & 1280 \\
        Heads & 3 & 6 & 10 & 20 \\
        Layers & 3 & 6 & 10 & 20 \\
        Feed Forward Dim & 576 & 1152 & 1920 & 3840 \\
        \midrule
        \multicolumn{5}{l}{\textit{Training}} \\
        Optimizer Steps & 1M & 2M & 3M & 4M \\
        Learning Rate $\eta_{\text{max}}$ & $10^{-4}$ & $10^{-4}$ & $10^{-4}$ & $3 \cdot 10^{-5}$ \\
        Weight Decay & 0.01 & 0.01 & 0.01 & 0.025 \\
        $\beta_2$ & 0.95 & 0.95 & 0.95 & 0.99995 \\
        Logical CPUs & 4 & 32 & 8 & 8 \\
        GPU & RTX 2080Ti & RTX 4090 & A100 & H200 \\
        Time Taken & 112h & 56h & 340h & 657h \\
        \midrule
        \multicolumn{5}{l}{\textit{Parameters}} \\
        Total & 3.2M & 23M & 122M & 955M \\
        \quad Encoder & 1.6M & 10.4M & 64.4M & 494M \\
        \quad Decoder & 1.6M & 12.4M & 57.5M & 459M \\
        \bottomrule
      \end{tabular}
    \end{small}
  \end{center}
  \vskip -0.1in
\end{table}

\paragraph{Simplification.}
We employ our \textsc{SimpliPy} symbolic simplification engine (version \texttt{dev\_7-3}) with a maximum pattern length of $L_{\max}=4$ during training and inference.

\section{Test-Time Compute System}
\label{app:test_time_compute_system}

All runtime-sensitive experiments for Section \ref{sec:results_test_time_compute} (Time-Normalized Recovery Rate) were conducted on a single representative high-end consumer workstation hardware setup (Table \ref{tab:test_time_compute_system}) to ensure comparability.

\begin{table}[ht]
  \caption{System specifications for Time-Normalized Recovery Rate experiment series.}
  \label{tab:test_time_compute_system}
  \begin{center}
    \begin{small}
      \begin{tabular}{ll}
        \toprule
        \textsc{Component} & \textsc{Specification} \\
        \midrule
        CPU & AMD Ryzen 9 9950X (16C, 32T) \\
        GPU & NVIDIA RTX 4090 (24 GB) \\
        RAM & 64 GB DDR5 6000MT/s \\
        \midrule
        OS & Ubuntu 24.04.3 LTS \\
        Python & 3.13.7 \\
        PyTorch & 2.9.1 + CUDA 13.0 \\
        \bottomrule
      \end{tabular}
    \end{small}
  \end{center}
  \vskip -0.1in
\end{table}

\section{Masked RMSNorm for Sets}
\label{app:masked_rmsnorm}
Let $Z \in \mathbb{R}^{N \times M \times d}$ be a tensor representing a batch of $N$ sets, each containing up to $M$ points with dimension $d$ padded with arbitrary values where necessary.
Let $\Xi \in \{0, 1\}^{N \times M}$ be a binary mask where $\Xi_{n,m} = 1$ indicates a valid point and $0$ indicates padding.
We introduce a learnable scaling parameter $\beta \in \mathbb{R}^d$.
For each set $n \in \{1, \dots, N\}$, we compute the masked root mean square statistic $\nu_n$ over all valid points and features:
\begin{equation}
\nu_n = \sqrt{ \frac{\sum_{m=1}^M \sum_{j=1}^d (Z_{n,m,j})^2 \cdot \Xi_{n,m}}{\max(1, d \cdot \sum_{m=1}^M \Xi_{n,m})} + \epsilon}
\end{equation}
where $\epsilon$ is a small constant for numerical stability.
The normalized output $\hat{Z}$ is then obtained by scaling the input element-wise:
\begin{equation}
\hat{Z}_{n,m,j} = \frac{Z_{n,m,j}}{\nu_n} \cdot \beta_j
\end{equation}

\newpage

\section{Model Architecture}
\label{app:model_architecture}

\begin{figure}[!ht]
  \begin{center}
    \centerline{\includegraphics[width=0.6\columnwidth]{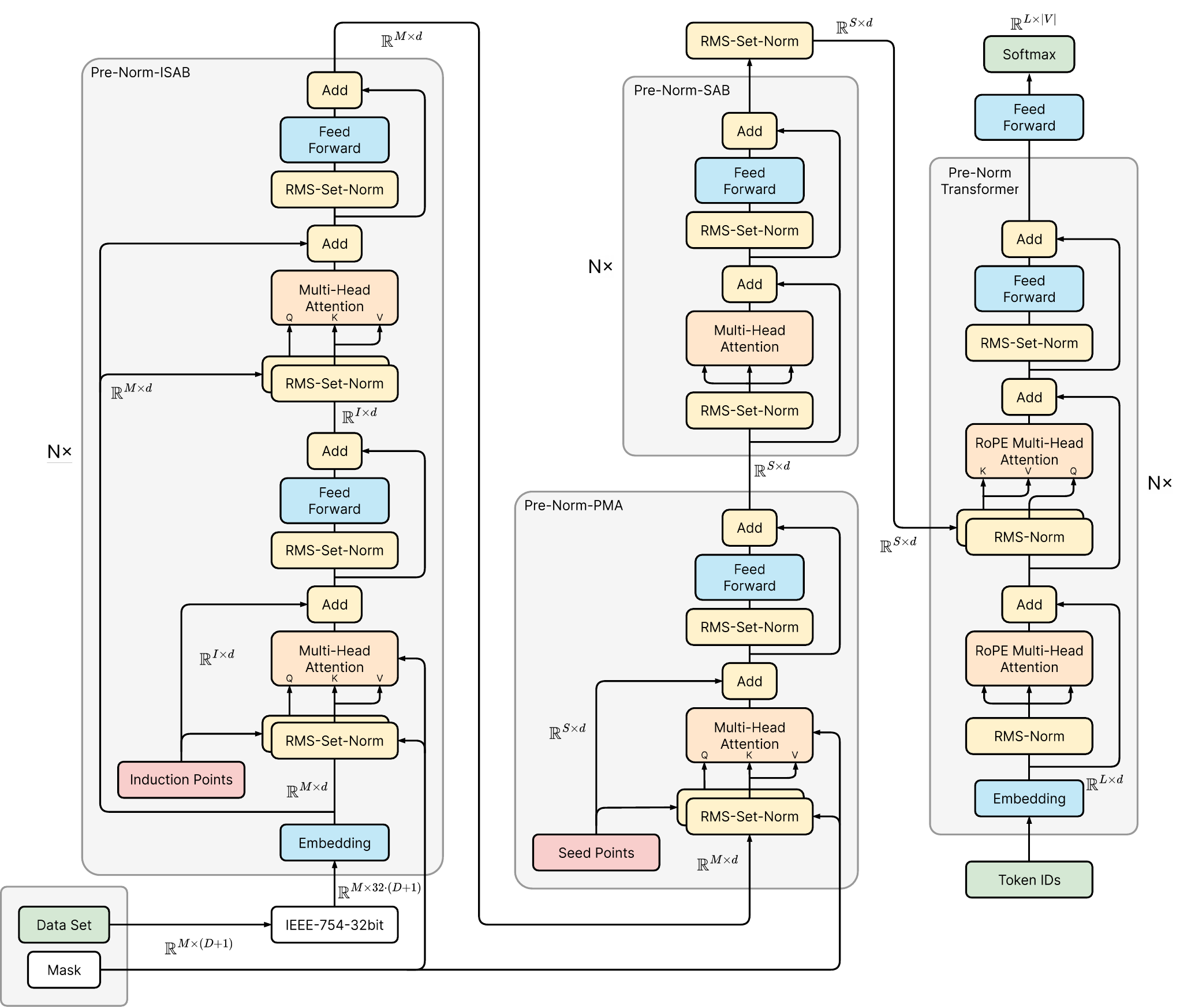}}
    \caption{
      Visual depiction of the Flash-ANSR model architecture. The Set Transformer encoder ingests a variable-sized set of input-output pairs and produces a fixed-size latent representation via Induced Set Attention Blocks (ISAB) and Set Attention Blocks (SAB). The Transformer decoder autoregressively generates a symbolic expression token-by-token, attending to the encoded dataset at each step.
    }
    \label{fig:model_architecture}
  \end{center}
  \vskip -0.3in
\end{figure}

\section{Operators}
\label{app:operators}

The set of operators supported by \textsc{Flash-ANSR} and \textsc{SimpliPy} is summarized in Table \ref{tab:vocabulary}.
During training, we sample \token{+}, \token{-}, \token{*}, and \token{/} with weight 10 and all others with weight 1.

\begin{table}[!ht]
  \caption{Flash-ANSR operator vocabulary.}
  \label{tab:vocabulary}
  \begin{center}
    \begin{small}
      \renewcommand{\arraystretch}{1.5}  
      \begin{tabular}{ll|ll|ll}
        \toprule
        \textsc{Token} & \textsc{Description} & \textsc{Token} & \textsc{Description} & \textsc{Token} & \textsc{Description} \\
        \midrule
        \token{+} & Addition & \token{pow1\_2} & $\sqrt{x}$ & \token{asinh} & $\operatorname{arcsinh}(x)$ \\
        \token{-} & Subtraction & \token{pow1\_3} & $\sqrt[3]{x}$ & \token{acosh} & $\operatorname{arccosh}(x)$ \\
        \token{*} & Multiplication & \token{pow1\_4} & $\sqrt[4]{x}$ & \token{atanh} & $\operatorname{arctanh}(x)$ \\
        \token{/} & Division & \token{pow1\_5} & $\sqrt[5]{x}$ & \token{exp} & $e^x$ \\
        \token{abs} & $|x|$ & \token{sin} & $\sin(x)$ & \token{log} & $\log(x)$ \\
        \token{inv} & $\frac{1}{x}$ & \token{cos} & $\cos(x)$ & \token{mult2} & $2 \cdot x$ \\
        \token{neg} & $-x$ & \token{tan} & $\tan(x)$ & \token{mult3} & $3 \cdot x$ \\
        \token{pow} & Exponentiation & \token{asin} & $\arcsin(x)$ & \token{mult4} & $4 \cdot x$ \\
        \token{pow2} & $x^2$ & \token{acos} & $\arccos(x)$ & \token{mult5} & $5 \cdot x$ \\
        \token{pow3} & $x^3$ & \token{atan} & $\arctan(x)$ & \token{div2} & $\frac{x}{2}$ \\
        \token{pow4} & $x^4$ & \token{sinh} & $\sinh(x)$ & \token{div3} & $\frac{x}{3}$ \\
        \token{pow5} & $x^5$ & \token{cosh} & $\cosh(x)$ & \token{div4} & $\frac{x}{4}$ \\
         &  & \token{tanh} & $\tanh(x)$ & \token{div5} & $\frac{x}{5}$ \\
        \bottomrule
      \end{tabular}
    \end{small}
  \end{center}
  \vskip -0.1in
\end{table}

\section{Algorithms}
\label{app:algorithms}

\camready{We provide formal versions of the algorithms used by \textsc{SimpliPy}, in two passes matching Section~\ref{sec:simplipy}'s phase split. We open with the offline rule-discovery procedure (Algorithm~\ref{alg:simplipy_rule_discovery}) and the numerical equivalence check (Algorithm~\ref{alg:equivalence}) it invokes, together with a discussion of $\mathcal{R}$'s non-minimality (Appendix~\ref{app:redundancy}); we then turn to the online subroutines \textsc{CancelTerms} (Algorithm~\ref{alg:cancel_terms}) and \textsc{ApplyRules} (Algorithm~\ref{alg:apply_rules}) that compose the abstract simplification loop of Algorithm~\ref{alg:simplipy_online} and their joint complexity analysis.}

\begin{algorithm}[!h]
\caption{SimpliPy Rule Discovery}
\label{alg:simplipy_rule_discovery}
\begin{algorithmic}[1]
\STATE \textbf{Input:} Operators $\mathcal{O}$ (Appendix \ref{app:operators}), Variables $\mathcal{V}$, Literals $\mathcal{C}$ (Section \ref{sec:simplipy}), Limits $L_{\max} = 7, L_{\text{rep}} = 3$, Data $X \sim \mathcal{N}(0, 5)^{1024 \times D}$, number of challenges $K = 16$ and retries $R = 16$
\STATE Initialize rule set $\mathcal{R} \leftarrow \emptyset$ and pre-compute $\mathcal{E}_{L}$, the set of expressions of length $L$ using $\mathcal{O}, \mathcal{V}, \mathcal{C}$ for $L = 1$ to $L_{\max}$
\FOR{length $L = 1$ to $L_{\max}$}
    \STATE Initialize temporary rule cache $\mathcal{R}_{\text{new}} \leftarrow \emptyset$
    \FOR{each expression $\boldsymbol{\tau} \in \mathcal{E}_{L}$}
        \STATE $\boldsymbol{\tau}^* \leftarrow \textsc{SimpliPy}_\mathcal{R}(\boldsymbol{\tau}); \quad \text{found} \leftarrow \text{False}; \quad \boldsymbol{\tau}'_{\text{best}} \leftarrow \text{None}$
        \IF{$|\boldsymbol{\tau}^*| < |\boldsymbol{\tau}|$}
            \STATE \textbf{continue} \hfill\COMMENT{Kruskal-style skip: $\boldsymbol{\tau}$ already reduces under $\mathcal{R}$; no new rule needed}
        \ENDIF
        \FOR{length $j = 1$ to $\min(|\boldsymbol{\tau}^*| - 1, L_{\text{rep}})$}
            \FOR{each $\boldsymbol{\tau}' \in \mathcal{E}_{j}$ where $\text{Vars}(\boldsymbol{\tau}') \subseteq \text{Vars}(\boldsymbol{\tau})$}
                \IF{$\text{Equivalent}(\boldsymbol{\tau}, \boldsymbol{\tau}'; X, K, R)$}
                    \IF{$\boldsymbol{\tau}'_{\text{best}} = \text{None}$ \textbf{or} $n_c(\boldsymbol{\tau}') < n_c(\boldsymbol{\tau}'_{\text{best}})$}
                        \STATE $\boldsymbol{\tau}'_{\text{best}} \leftarrow \boldsymbol{\tau}'$
                    \ENDIF
                    \STATE $\text{found} \leftarrow \text{True}$
                \ENDIF
            \ENDFOR
            \IF{$\text{found}$ \textbf{and} no metavariable in $\boldsymbol{\tau}'_{\text{best}}$ occurs more often than in $\boldsymbol{\tau}$}
                \STATE $\mathcal{R}_{\text{new}} \leftarrow \mathcal{R}_{\text{new}} \cup \{ \boldsymbol{\tau} \to \boldsymbol{\tau}'_{\text{best}} \}$; \textbf{break}
            \ENDIF
        \ENDFOR
    \ENDFOR
    \STATE $\mathcal{R} \leftarrow \mathcal{R} \cup \textsc{Deduplicate}(\mathcal{R}_{\text{new}})$ \hfill\COMMENT{under variable-canonicalization}
    \STATE Rehash explicit rules into $\mathcal{R}_g$; rebucket pattern rules into $\mathcal{R}_v$ by $(\textsc{root}, \text{length})$
\ENDFOR
\STATE \textbf{return} $\mathcal{R}$
\end{algorithmic}
\end{algorithm}

\camready{
\subsection{Design Choices in Rule Discovery}
\label{app:redundancy}

The discovery procedure of Algorithm~\ref{alg:simplipy_rule_discovery} fixes $\mathcal{R}$ for the duration of each length-$L$ pass: new rules accumulate in a side cache $\mathcal{R}_{\text{new}}$ and are merged into $\mathcal{R}$ only after the pass completes. This makes the inner loop embarrassingly parallel (every $\bm{\tau} \in \mathcal{E}_L$ is tested against the same snapshot of $\mathcal{R}$, with no inter-worker coordination) and is what allows the 32-thread discovery run to complete in ${\sim}100\,$h. In exchange, distinct expressions in $\mathcal{E}_L$ cannot see rules just admitted within the same pass, so a fraction of the rules in the published $\mathcal{R}$ are subsumed by other rules from the same length stratum and could be removed without changing $\textsc{SimpliPy}_\mathcal{R}$ on any input.

A minimal example illustrates the situation. At $L = 3$, $\bm{\tau}_W = x_0 \cdot 0$ and $\bm{\tau}_E = 1 \cdot 0$ are both irreducible under $\mathcal{R}$ at the start of the pass: \textsc{CancelTerms} counts $\textsc{add}$/$\textsc{mult}$ multiplicities but does not encode zero as a multiplicative annihilator. The inner search finds $\bm{\tau}' = 0$ equivalent to each, so the pass admits the pattern rule $W: \square_1 \cdot 0 \to 0$ from $\bm{\tau}_W$ (which carries a variable) and the explicit rule $E: 1 \cdot 0 \to 0$ from $\bm{\tau}_E$ (which does not). After commit, $W$ matches $t \cdot 0$ for any subterm $t$, including $t = 1$; removing $E$ leaves $\textsc{SimpliPy}_\mathcal{R}$ unchanged.

The redundancy is strictly intra-length. A length-$L_2$ metapattern with $L_2 > L_1$ requires a matched subterm of length $\geq L_2$ and so cannot subsume any length-$L_1$ rule, and when a length-$L_2$ rule is admitted, every shorter rule is already in $\mathcal{R}$ and has been applied in $\textsc{SimpliPy}_\mathcal{R}(\bm{\tau})$ before any new rule is considered. Same-length subsumption forces the substitution between the two metapatterns to be length-preserving, which collapses each metavariable to a single token as illustrated.

A removal-and-recheck pruning pass would eliminate the residual redundancy at $O(R \cdot T_{\textsc{SimpliPy}})$ additional time.
The deployment cost of the extra rules is small: the online complexity bound $O(K \cdot n^2 \cdot R \cdot P + n^2 \log n)$ absorbs them linearly in $R$, and bucketing of $\mathcal{R}_v$ by root operator and pattern length further dampens the constant in practice.
}

\begin{algorithm}[t]
\caption{Expression Equivalence Check}
\label{alg:equivalence}
\begin{algorithmic}[1]
\STATE \textbf{Input:} pattern $\boldsymbol{\tau}$, candidate replacement $\boldsymbol{\tau}'$, data $X$, challenges $K = 16$, retries $R = 16$, tolerance $\epsilon$ (caller-specified)
\STATE Let $C$ be constants of $\boldsymbol{\tau}$; $C'$ be constants of $\boldsymbol{\tau}'$
\STATE \textbf{Convention:} $\text{allclose}(a, b, \epsilon)$ treats NaN entries as matches (\texttt{equal\_nan=True}); the conservative bias from this convention is what makes the caller's decontamination filter err on the side of rejection.
\FOR{$k = 1$ to $K$}
    \STATE Sample $r \sim \mathcal{N}(0,\sigma=5)^{|C|}$
    \FOR{$s \in \{-1,0,1\}^{|C|}$}
        \STATE $y_{\boldsymbol{\tau}} \leftarrow f_{\boldsymbol{\tau}}(X; |r| \odot s)$
        \IF{$|C'| = 0$}
            \STATE $\text{match} \leftarrow \text{allclose}(f_{\boldsymbol{\tau}'}(X), y_{\boldsymbol{\tau}}, \epsilon)$
        \ELSE
            \STATE $\text{match} \leftarrow \text{False}$
            \FOR{$j = 1$ to $R$}
                \STATE $p_0 \sim \mathcal{N}(0,\sigma=5)^{|C'|}$
                \IF{$\exists \hat{c} \text{ s.t. } \text{allclose}(f_{\hat{\boldsymbol{\tau}}}(X; \hat{c}), y_{\boldsymbol{\tau}}, \epsilon)$ with Levenberg--Marquardt initialized at $p_0$}
                    \STATE $\text{match} \leftarrow \text{True}$; \textbf{break}
                \ENDIF
            \ENDFOR
        \ENDIF
        \IF{$\neg \text{match}$}
          \STATE \textbf{return} False
        \ENDIF
    \ENDFOR
\ENDFOR
\STATE \textbf{return} True
\end{algorithmic}
\end{algorithm}

\newpage

\camready{
\begin{algorithm}[tb]
\caption{\textsc{CancelTerms}: multiplicity-based cancellation in additive and multiplicative subtrees.}
\label{alg:cancel_terms}
\begin{algorithmic}[1]
\STATE \textbf{Input:} expression $\boldsymbol{\tau}$ in prefix notation
\STATE $T \gets \textsc{ParseTree}(\boldsymbol{\tau})$
\STATE Classes: $\textsc{add} = (\{+, -\},\; \text{neutral } 0,\; \text{hyperoperator } \textsc{mult})$;\, $\textsc{mult} = (\{*, /\},\; \text{neutral } 1,\; \text{hyperoperator } \textsc{pow})$. Inverse operators ($-$, $/$) flip the polarity of their right operand.
\STATE \emph{Pass 1. bottom-up multiplicity annotation:}
\FOR{each subtree $s$ in $T$, visited bottom-up}
    \FOR{each class $\mathcal{C} \in \{\textsc{add}, \textsc{mult}\}$}
        \STATE $(n_s^{\mathcal{C}, +}, n_s^{\mathcal{C}, -}) \gets$ multiplicities of $s$ as a positive / negative operand within the enclosing class-$\mathcal{C}$ chain (counted across nested operators of the same class; inverse-operator polarity flips applied at right operands)
    \ENDFOR
\ENDFOR
\STATE \emph{Pass 2. select cancellation target + rewrite:}
\STATE Pick the first $(s^*, \mathcal{C}^*)$ in bottom-up order with $n_{s^*}^{\mathcal{C}^*, +} + n_{s^*}^{\mathcal{C}^*, -} > 1$, skipping composite operands containing the constant placeholder $\diamond$.
\STATE Let $m \gets n_{s^*}^{\mathcal{C}^*, +} - n_{s^*}^{\mathcal{C}^*, -}$.
\IF{$m = 0$}
    \STATE replace every occurrence of $s^*$ in the $\mathcal{C}^*$-chain with the class-neutral element
\ELSIF{$|m| = 1$}
    \STATE keep the first occurrence (with sign $\mathrm{sgn}(m)$); replace the rest with the neutral element
\ELSE
    \STATE replace the first occurrence with $\text{hyperoperator}_{|m|}(s^*)$, prefixed by the class-inverse operator iff $\mathrm{sgn}(m) < 0$; replace the rest with the neutral element \hfill\COMMENT{if $|m|$ exceeds the available hyperoperator token range, fall back to a binary op with a numeric coefficient/exponent}
\ENDIF
\STATE \textbf{return} $\textsc{Flatten}(T)$
\end{algorithmic}
\end{algorithm}
}

\newpage

\camready{
\begin{algorithm}[!h]
\caption{\textsc{ApplyRules}: pattern matching against the discovered rule set.}
\label{alg:apply_rules}
\begin{algorithmic}[1]
\STATE \textbf{Input:} expression $\boldsymbol{\tau}$, explicit rules $\mathcal{R}_g$ (hash-indexed by prefix sequence), pattern rules $\mathcal{R}_v$ (bucketed by key $(\textsc{root}(t), |p|)$; within each bucket, rules are iterated in decreasing $|p|$)
\STATE $T \gets \textsc{ParseTree}(\boldsymbol{\tau});\quad T \gets \textsc{Rewrite}(T)$
\STATE \textbf{return} $\textsc{Flatten}(T)$
\STATE \hrulefill
\STATE \textbf{subroutine} $\textsc{Rewrite}(t)$:
\IF{$t$ is a leaf}
    \STATE \textbf{return} $t$
\ENDIF
\IF{every operand of $t$ is the constant placeholder \token{$\diamond$}}
    \STATE \textbf{return} \token{$\diamond$} \hfill\COMMENT{constant folding}
\ENDIF
\STATE \emph{Pass 1. try to rewrite the current node:}
\IF{$\exists r : (t \to r) \in \mathcal{R}_g$}
    \STATE \textbf{return} $\textsc{Rewrite}(r)$
\ENDIF
\FOR{rules $(p \to r) \in \mathcal{R}_v[(\textsc{root}(t), |p|)]$ in decreasing $|p|$}
    \IF{a substitution $\sigma$ exists with $p[\sigma] = t$ (syntactic matching on the prefix representation)}
        \STATE \textbf{return} $\textsc{Rewrite}(r[\sigma])$
    \ENDIF
\ENDFOR
\STATE \emph{Pass 2. recurse into children, then retry:}
\FOR{each child $c$ of $t$}
    \STATE $c \gets \textsc{Rewrite}(c)$
\ENDFOR
\STATE let $t'$ be the resulting node; retry Pass 1 on $t'$ (explicit lookup, then pattern bucket)
\STATE \textbf{return} $t'$
\end{algorithmic}
\end{algorithm}
}

\camready{
\subsection{Complexity Analysis}
We derive the worst-case complexity of online simplification (Algorithm~\ref{alg:simplipy_online} including Algorithms \ref{alg:cancel_terms}, \ref{alg:apply_rules} as subroutines) for an input expression with $n$ tokens, a rule set $\mathcal{R}$ of $R$ rules with maximum pattern length $P$, and the iteration cap $K$.

\textbf{Subroutine 1: Cancellation (Algorithm~\ref{alg:cancel_terms} above).} Cancellation runs in two stack-based passes over the prefix tree. The bottom-up multiplicity pass visits each token once, maintaining per-class multiplicity tables on a stack of subtree annotations. Merging at each binary node touches only that node's operands, so each subtree contributes a constant amount of work along its root-to-leaf path. The top-down rewrite pass selects subtrees with $n^+ + n^- > 1$ and rewrites them, touching a bounded number of operands per subtree. Thus, the combined cost is $O(n)$ per call.

\textbf{Subroutine 2: Rule application (Algorithm~\ref{alg:apply_rules} below).}
Per node, \textsc{Rewrite} performs (i) a constant-time hash lookup in explicit rules $\mathcal{R}_g$, then (ii) iterates over the pattern rule buckets $\mathcal{R}_v[(\textsc{root}(t), |p|)]$ for $|p|$ from $P$ down to $1$, with each match attempt costing $O(P)$ for syntactic matching and substitution. The worst-case bucket size is $R$ (loose bound), so per-node matching is $O(R \cdot P)$. Without rule matches, the recursion visits each of the $m$ nodes in the input subtree a constant number of times (initial Pass-1 attempt and post-children retry), giving $O(m \cdot R \cdot P)$.

When rules match, each application strictly reduces token length by the size-decreasing property of Section~\ref{sec:simplipy}, and the substituted replacement is recursively rewritten by a fresh \textsc{Rewrite} call. Let the chain of recursive calls produced by a single top-level invocation have subtree sizes $m = m_0 > m_1 > \cdots > m_q \ge 1$. The total cost of the chain is
\begin{equation*}
\sum_{k=0}^{q} m_k \cdot R \cdot P \;\le\; R \cdot P \cdot \sum_{k=0}^{n-1}(n - k) \;=\; O(n^2 \cdot R \cdot P).
\end{equation*}
This bound is achieved when each rewrite reduces length by exactly one and the entire shrinking expression is re-traversed by the next \textsc{Rewrite} call.

\newpage

\textbf{Iteration cap.} The outer loop of Algorithm~\ref{alg:simplipy_online} runs at most $K$ times, with early termination on convergence. Across $K$ iterations, cancellation and rule application accumulate to $O(K \cdot n^2 \cdot R \cdot P)$ (rule application dominates).

\textbf{Operand sorting.} The final canonicalization pass runs once after the iteration loop. It sorts commutative-operator chains lexicographically by a recursive structural key. With up to $n$ comparable operands in a single chain, each pairwise key comparison costs $O(n)$ for full subtree traversal in the absence of cached hashes, and a sort makes $O(n \log n)$ comparisons. Total: $O(n^2 \log n)$.

In total, the worst-case complexity of online simplification is
\begin{equation}
\label{eq:online_complexity}
T_\text{online}(n, R, P, K) \;=\; O\bigl(K \cdot n^2 \cdot R \cdot P \,+\, n^2 \log n\bigr).
\end{equation}
The first term captures rule-driven simplification across $K$ iterations, the second the one-time canonical-ordering pass.

\textbf{Practical Notes.} The $n^2 \log n$ term is tight for fully commutative expressions (e.g. a single $+$ chain over $n$ leaves). The $K \cdot n^2 \cdot R \cdot P$ term is tight when rule applications cascade and the entire shrinking expression is re-traversed by the recursive \textsc{Rewrite} call. Bucketing of $\mathcal{R}_v$ by $(\textsc{root}(t), |p|)$ does not improve the worst-case bound (a single bucket can in principle contain all $R$ rules) but reduces the practical constant: rules distribute across $O(P \cdot |\mathcal{O}|)$ buckets, so observed bucket sizes are far below $R$. Most expressions converge in 1--3 iterations, well below the $K = 5$ cap.}

\section{Baselines}
\label{app:baselines}
Baseline methods are configured at or as close to their defaults as possible while matching the problem setting.

For PySR, we allow the same operator set as \textsc{Flash-ANSR}, excluding explicit integer multiplication and division operators.

For NeSymReS, we evaluate the ``100M'' model trained on 100M expressions with the default 4 optimizer restarts.

For E2E, we employ the released \texttt{model1.pt} with default refinement parameters (10 trees, 100 bags, 200 points per bag).
To attempt evaluation at higher inference budgets (beam width $\ge 512$), we apply a dynamic reduction of the maximum generation length to satisfy internal batching constraints.
We could not get E2E to function beyond a beam width of 512.

For both, we apply \emph{minimal} compatibility patches to the official implementation to ensure basic functionality in the evaluation environment.
These patches are documented and made publicly available alongside our codebase.

\section{Excluded FastSRB Equations}
\label{app:excluded_equations}

\camready{We exclude 5 of the 120 \textsc{FastSRB} equations (B4, B7, B17, II.24.17, III.14.14) whose prescribed sampling regimes induce \texttt{sqrt}-domain or \texttt{exp}-overflow issues that the benchmark's $100{\times}$ resampling protocol fails to resolve.
Concretely, the prescribed input domains for these equations either include negative values inside a \texttt{sqrt} or exponents large enough to overflow 32-bit floats with high probability, and resampling produces the same family of issues across attempts.
We treat this as a property of the SRSD/\textsc{FastSRB} sampling specification rather than of the SR methods evaluated against it: every method we evaluate fails uniformly on these problems for the same data-domain reason, regardless of inference budget.}

\section{Data Generation Hyperparameters}
\label{app:data_gen_example}

\camready{We list the priors and hyperparameters of the on-the-fly data-generation pipeline introduced in Section~\ref{sec:data_generation}.}

\camready{
\subsection{Skeleton Sampling}
\begin{itemize}
\item Operator count: $n_\text{ops} \sim p_\text{ops}(n_\text{ops}) \propto \exp[n_\text{ops}^\alpha / \lambda]$ with $\alpha = 0.7$, $\lambda = 1$, support $n_\text{ops} \in \{0, 1, \dots, 17\}$.
\item Maximum skeleton length: $35$ symbols.
\item Operator weights: $\{$\token{+},\, \token{-},\, \token{*},\, \token{/}$\}$ at relative weight $10$, with all other operators (Appendix~\ref{app:operators}) at weight $1$.
\item Leaf cardinality: number of unique variables $n_{\text{var}^*} \sim \mathcal{U}(1, \min(n_\text{leaves}, D))$, where $n_\text{leaves}$ is the number of leaves in the prefix tree and $D$ is the considered maximum dimensionality.
\item Variable subset: drawn uniformly without replacement from $\mathcal{V} \cup \{\diamond\}$, then duplicated and shuffled to fill $n_\text{leaves}$ positions.
\end{itemize}

\subsection{Simplification}
\textsc{SimpliPy} is applied to the sampled skeletons with $L_\text{max} = 4$. Those containing non-finite tokens after simplification (\texttt{float("inf")}, \texttt{float("-inf")}, \texttt{float("nan")}) are rejected.

\subsection{Decontamination}
Constants are stripped from both candidate and held-out skeletons before comparison. Symbolic equality of pruned forms triggers immediate rejection. Otherwise, both skeletons are evaluated on a fixed grid $X_\text{check} \sim \mathcal{U}(-10, 10)^{512 \times D}$ sampled once per worker at startup, the resulting images are rounded element-wise to four decimal places, and a candidate is rejected on hash collision with any held-out image. NaN entries are replaced with $0$ before hashing, biasing the filter toward rejection at NaN or near-zero positions.

\subsection{Dataset Rendering}
\begin{itemize}
\item Number of data points: $M \sim \mathcal{U}(1, 1024)$.
\item Per-dimension support range: $a_j, b_j \sim \mathcal{N}(0, \sigma = 10)$ truncated to $[-30, 30]$.
\item Per-dimension data points: $X_j \sim \mathcal{U}(a_j, b_j)$.
\item Constants: $c_k \sim \mathcal{N}(0, \sigma = 5)$ i.i.d. per constant.
\item Validity guard: instances yielding complex or non-finite $y = f_{\bm{\bar{\tau}}}(X; \bm{c})$ are resampled from the same skeleton up to four times before the entire instance is rejected.
\end{itemize}

Statistics of the resulting training distribution are in Appendix~\ref{app:data_statistics}.
}

\newpage

\section{Data Statistics}
\label{app:data_statistics}
\begin{figure}[!h]
  \begin{center}
    \centerline{\includegraphics[width=0.5\columnwidth]{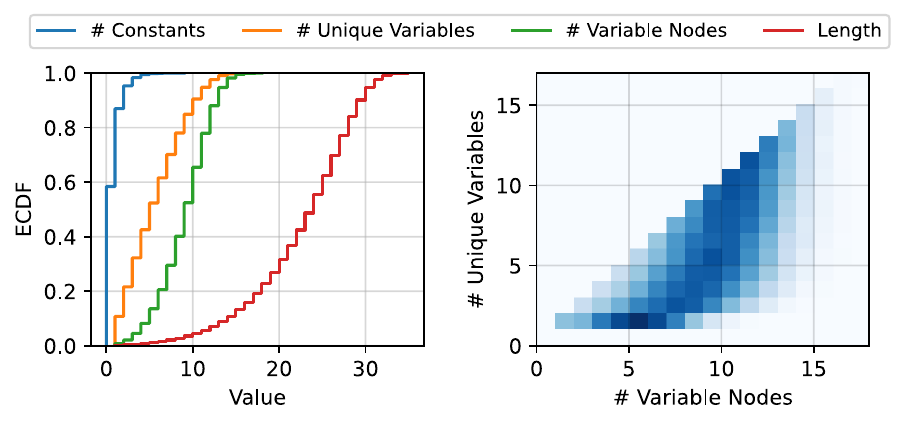}}
    \caption{
        Training data statistics. Left: Empirical Cumulative Distribution Functions (ECDFs) of the number of constants, unique variables, variable nodes and the expression length in the training data. Half of the simplified training expressions contain at least 5 unique variables and 24 symbols. Right: 2D histogram showing the number of unique variables and the number of variable nodes. The training expressions cover both low and high-dimensional regimes with complexity arising both due to the coupling of like and unlike variables.
    }
    \label{fig:additional_data_histograms}
    \vspace{-10pt}
  \end{center}
\end{figure}

\section{Training Dynamics and Scaling}
\label{app:training_dynamics}

\begin{figure}[!h]
  \begin{center}
    \centerline{\includegraphics[width=0.55\columnwidth]{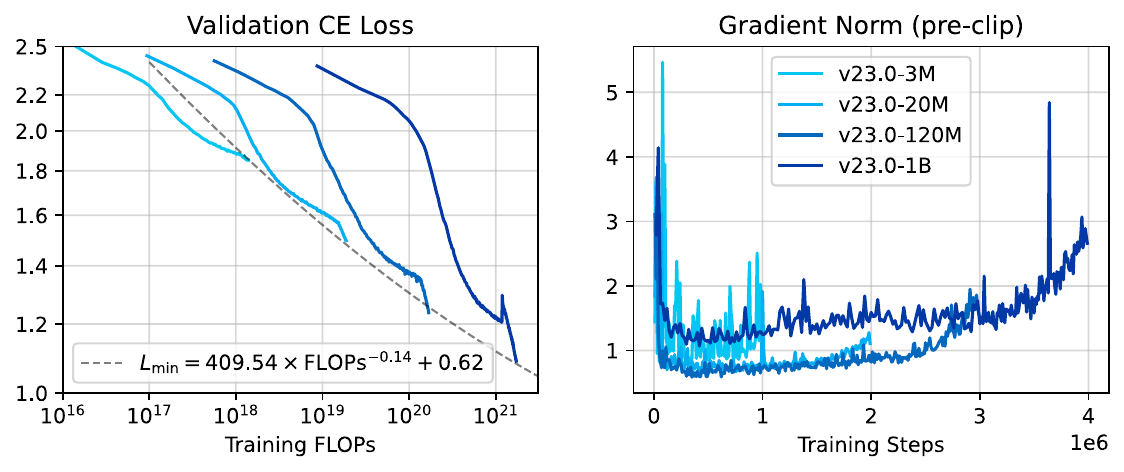}}
    \caption{
      Training dynamics and scaling behavior.
      \textbf{Left:} Validation Cross-Entropy loss (log-scale) as a function of estimated training FLOPs (log-scale). We use the approximation $\text{FLOPs}=6ND$ following \citet{kaplan2020scalinglawsneurallanguage}, and count FLOPs separately for the encoder with its dataset input and the decoder with its tokenized expression input before adding both for the total FLOPs.
      The performance envelope generally follows a power-law scaling relationship ($L \propto \text{FLOPs}^{-0.14}$).
      \textbf{Right:} Evolution of gradient norms (pre-clipping).
      Optimization is stable across the 3M--120M regime. The 1B model exhibits one gradient spike late in training from which it fully recovers over the subsequent training steps. Further investigation is required to determine if this is a systemic issue at large model sizes.
    }
    \label{fig:training_curves}
  \end{center}
\end{figure}

\begin{figure}[!h]
  \begin{center}
    \centerline{\includegraphics[width=0.55\columnwidth]{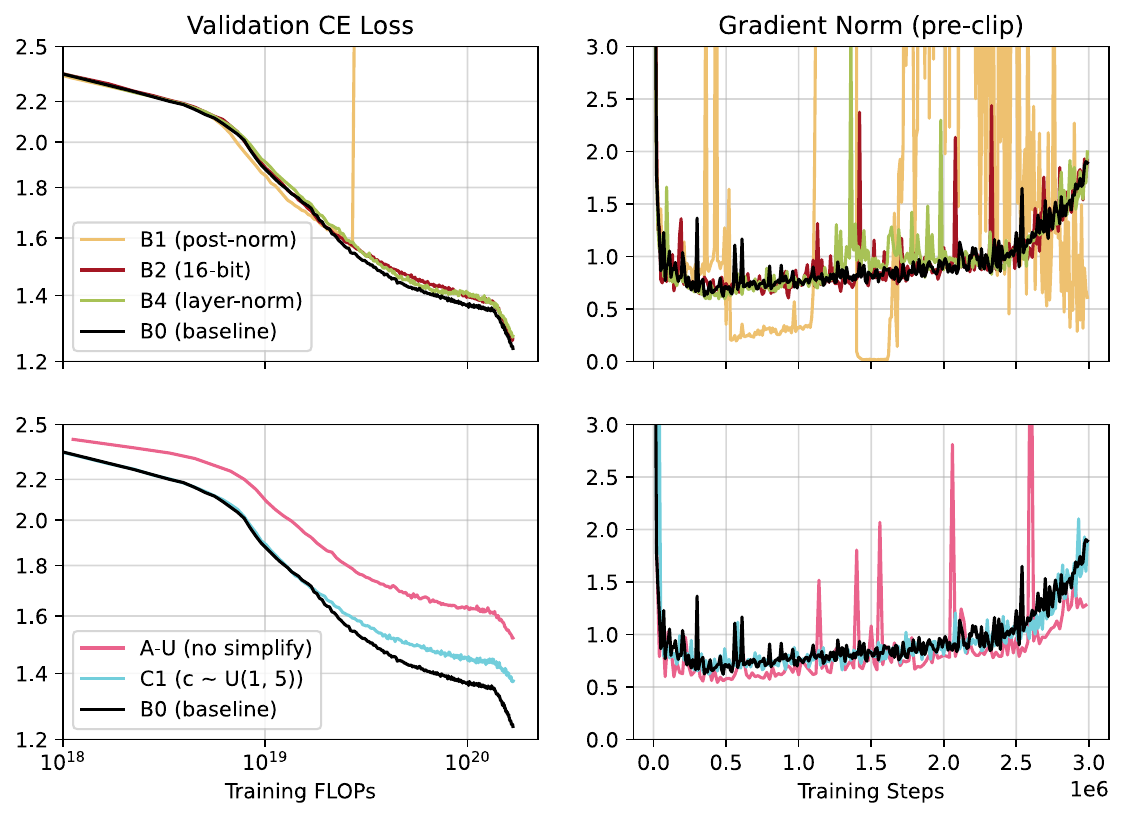}}
    \caption{
      \camready{Training dynamics of the system-effect ablations against \textit{B0} (black, shown in both rows), a configuration-matched retrain of \textsc{v23.0-120M}.
      \textbf{Top:} Architecture axis (B-track). \textit{B1} (Post-Norm decoder), \textit{B2} (16-bit input encoding), \textit{B3} (LayerNorm encoder).
      \textbf{Bottom:} Training-data axis (A-track / C-track). \textit{A-U} (unsimplified training expressions), \textit{C1} (narrow $\mathcal{U}(1,5)$ constants prior).
      \textbf{Left:} Validation Cross-Entropy loss vs.\ estimated training FLOPs.
      \textbf{Right:} Gradient norms (pre-clipping) vs.\ training steps.
      B1 (Post-Norm) shows a substantial loss spike around $2\times10^{19}$ FLOPs and sustained gradient instability (alternating between vanishing-gradient regions and large spikes from ${\sim}500$k steps onward).
      B2 (16-bit input encoding) and B3 (LayerNorm encoder) track the baseline more closely with only occasional gradient spikes.
      The training-data ablations exhibit stable optimization but converge to higher loss floors than the baseline (A-U most pronounced).
      We list detailed per-ablation metrics in Appendix~\ref{app:ablations}.}
    }
    \label{fig:training_curves_ablations_B}
  \end{center}
\end{figure}

\newpage

\section{Additional Metrics}
\label{app:metrics}

We use the following definition for the Fraction of Variance Unexplained (FVU):
\begin{equation}
  \label{eq:fvu}
    \text{FVU}(y, \hat{y}) = \frac{\sum_{i=1}^N (y_i - \hat{y}_i)^2}{\sum_{i=1}^N (y_i - \bar{y})^2}
\end{equation}
with $\bar{y} = \frac{1}{N} \sum_{i=1}^N y_i$.

Beyond the primary results, we report the following diagnostics with 95\% confidence intervals derived from $1000$-fold bootstrapping.

\textbf{Log$_{10}$ FVU (non-perfect fits).}
For cases that are \emph{not} numerically perfect ($\mathrm{FVU}>\epsilon_{32}$), we summarize $\log_{10}(\mathrm{FVU})$.
Exact fits are excluded before aggregation.

\textbf{Numeric Recovery Rate (NRR).}
For a benchmark with $N_\text{test}$ problems, NRR is the fraction of problems where the predicted expression achieves a numerically perfect fit on the data points $y^{(i)} \in \mathbb{R}^{M\times D}$ of each problem $i$:
\begin{equation}
    \text{NRR} = \frac{1}{N_\text{test}} \sum_{i=1}^{N_\text{test}} \mathbb{I}\left (\text{FVU}(y^{(i)}, \hat{y}^{(i)}) \le \epsilon_{32} \right )
\end{equation}
with $\epsilon_{32} \approx 1.19 \times 10^{-7}$ being the machine epsilon for 32-bit floating point numbers.
\camready{We report fNRR and vNRR for the fit split and an unseen validation split of equal size.}

\textbf{Symbolic Recovery Rate (SRR).}
Exact structural match between simplified predicted and ground-truth prefix skeletons:
\begin{equation}
\text{SRR} = \frac{1}{N_\text{test}} \sum_{i=1}^{N_\text{test}} \mathbb{I}\!\left(\textsc{SimpliPy}\left(\bm{\hat{\bar{\tau}}}^{(i)}\right) = \textsc{SimpliPy}\left(\bm{\bar{\tau}}^{(i)}\right)\right).
\end{equation}

\textbf{Expression F1.}
Token-level F1 between the predicted prefix skeleton and the simplified ground-truth skeleton.

\textbf{ZSS Tree Edit Distance.}
Zhang--Shasha tree edit distance \cite{zhang1989simple} between predicted and ground-truth skeleton trees.

\textbf{Variable Identification.}
Recall on the \emph{set of unique variables} present in the simplified predicted vs. ground-truth expressions.

\textbf{Log Probability.}
Mean log-probability of the generated sequence under the model.

\textbf{Excess Constants.}
Difference in the number of constants between prediction and ground truth. Positive values indicate an overuse of constants to fit the data.

\textbf{Expression Length Ratio.}
Length of the predicted prefix skeleton divided by the length of the simplified ground-truth skeleton.

\textbf{Total Nestedness.}
Sum of consecutive nestings contributed by unary operators over the predicted expression as a proxy for naturalness and interpretability.
For example, $y = \sin(\log(\operatorname{arccosh}(x)))$ has a total nestedness of $2$, and $y = \sin(\log(x_1 + \diamond))$ has a total nestedness of $1$.

\camready{%
\textbf{Rewrite Fraction.}
For each test problem the model returns $K$ candidate predictions $\{\bm{\hat{\tau}}_k\}_{k=1}^{K}$. Two candidates are \emph{rewrite-equivalent} if their \textsc{SimpliPy}-normalized prefix skeletons are identical. The Rewrite Fraction is the share of candidates that collide under this normalization, averaged across problems:
\begin{equation}
  \mathrm{Rewrites} = \mathbb{E}_{\text{problem}}\!\left[\,1 \;-\; \frac{\bigl|\{\textsc{SimpliPy}(\bm{\hat{\tau}}_k) : k = 1, \dots, K\}\bigr|}{K}\,\right].
\end{equation}
Higher values indicate rewrite-mode collapse: the budget is spent on algebraic restatements of the same canonical form rather than on functionally distinct hypotheses.

\textbf{Bloat.}
For a candidate $\bm{\hat{\tau}}$ with \textsc{SimpliPy}-normalized form $\textsc{SimpliPy}(\bm{\hat{\tau}})$, Bloat is the average relative inflation of raw outputs over their simplified equivalents, minus one (so canonical outputs score $0$):
\begin{equation}
  \mathrm{Bloat} \;=\; \mathbb{E}_{\bm{\hat{\tau}}}\!\left[\,\frac{|\bm{\hat{\tau}}|}{|\textsc{SimpliPy}(\bm{\hat{\tau}})|}\,\right] - 1.
\end{equation}
A value of $50\%$ corresponds to raw predictions $1.5\times$ as long as their canonical forms on average. Unlike Expression Length Ratio, Bloat requires no ground-truth reference: it is computed entirely from the model's outputs and the simplification engine.
}

\textbf{Relative Fit Time.}
Fit-time multiplier relative to the hyperparameter configuration for the test-time compute experiment.

\textbf{Fit Success.}
Fraction of test cases where the fitting procedure completed without errors.

\newpage

\subsection{Detailed Test Time Scaling Results}
\label{app:test_time_scaling}

\begin{figure}[!ht]
  \begin{center}
    \centerline{\includegraphics[width=0.8\columnwidth]{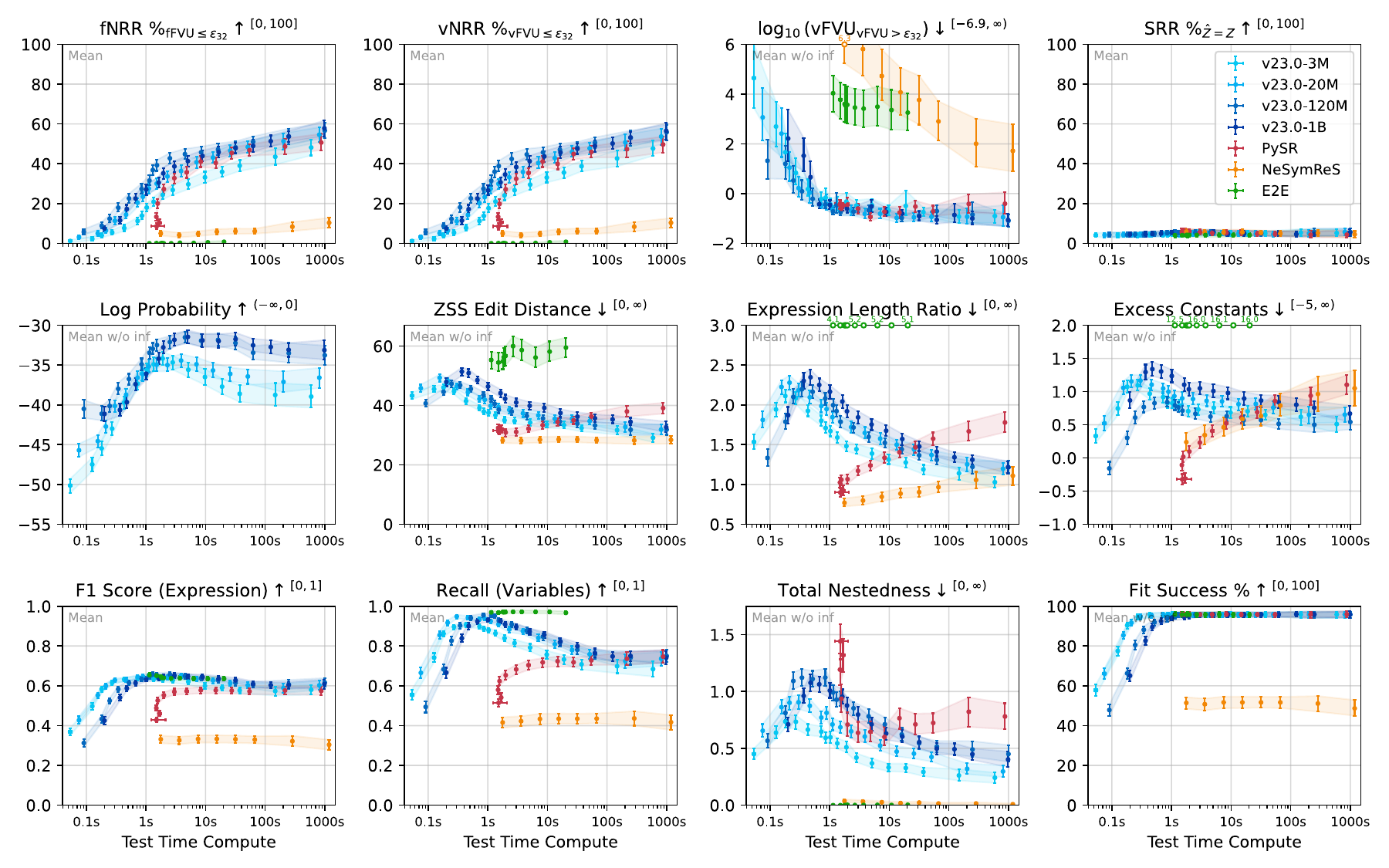}}
    \caption{
        Detailed comparison of time-normalized numerical and symbolic metrics.
        With increasing time budget, \textsc{Flash-ANSR} consistently generates more concise (lower expression length ratio), specific (fewer excess constants), and natural (lower total nestedness) expressions than PySR.
    }
    \label{fig:test_time_compute_fastsrb}
  \end{center}
    \vspace{-20pt}
\end{figure}

\subsection{Detailed Data Sparsity Results}
\label{app:data_sparsity}

\begin{figure}[!ht]
  \vskip 0.2in
  \begin{center}
    \centerline{\includegraphics[width=0.8\columnwidth]{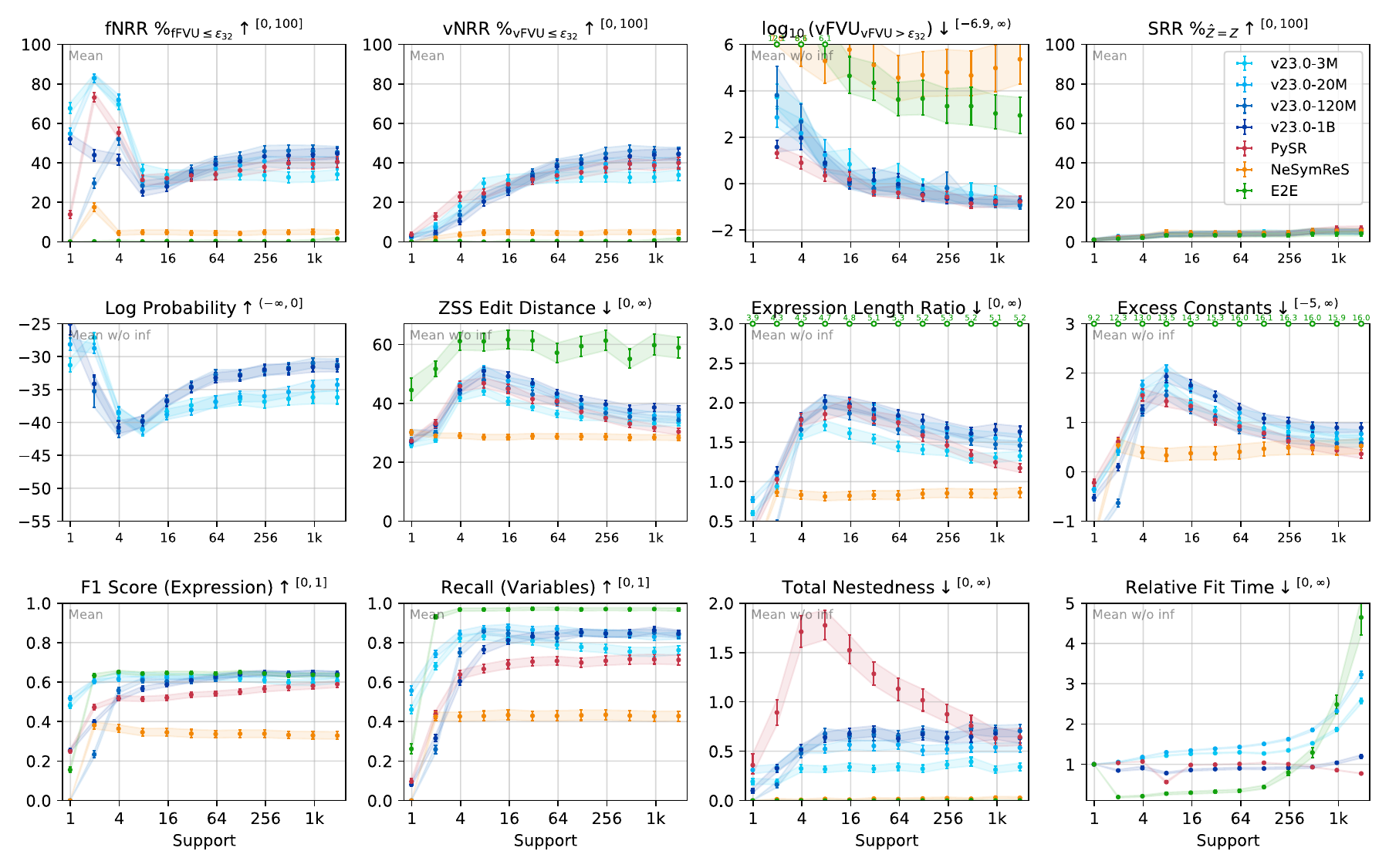}}
    \caption{
      Detailed results for varying number of data points $M$ on the \textsc{FastSRB} benchmark.
      In the Interpolation Regime ($N \approx 16$), \textsc{Flash-ANSR} maintains a lower mean total nestedness than PySR.
    }
    \label{fig:n_support_fastsrb}
  \end{center}
\end{figure}

\newpage

\subsection{Detailed Noise Level Scaling Results}
\label{app:noise_level_scaling}

\begin{figure}[ht]
  \vskip 0.2in
  \begin{center}
    \centerline{\includegraphics[width=0.8\columnwidth]{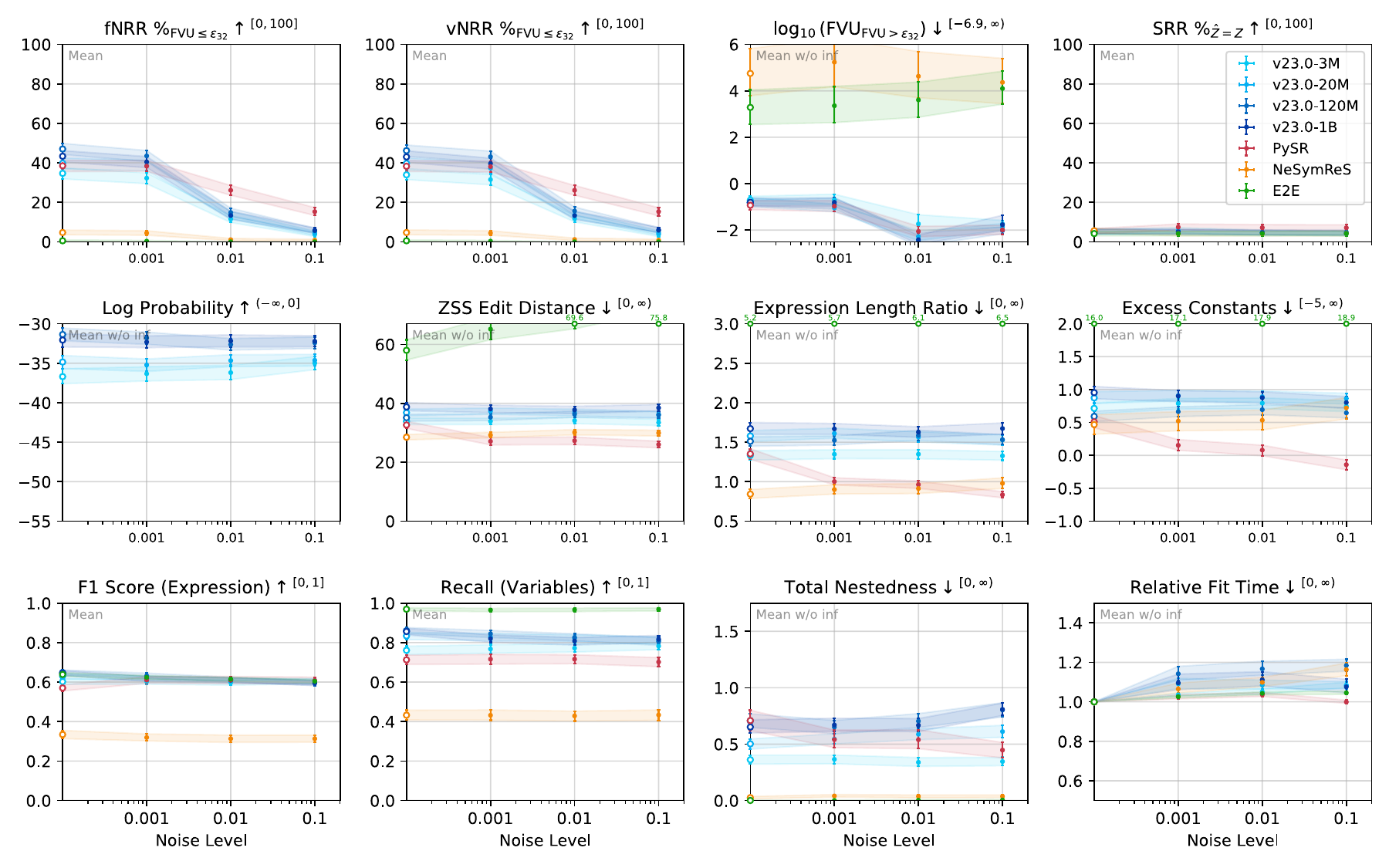}}
    \caption{
      Detailed results for varying noise levels $\eta$ on the \textsc{FastSRB} benchmark.
      Across all noise levels, \textsc{Flash-ANSR}'s expressions maintain similar quality.
      With higher noise levels, PySR still achieves shorter and more specific expressions.
    }
    \label{fig:noise_scaling_fastsrb}
  \end{center}
\end{figure}

\camready{
\section{Ablations}
\label{app:ablations}

We ablate four axes of the deployed \textsc{v23.0-120M} model:
\begin{itemize}
\item \textbf{Training data}: \textsc{A-U} (no \textsc{SimpliPy} on training expressions) and \textsc{C1} (narrow $\mathcal{U}(1,5)$ constants prior).
\item \textbf{Architecture}: \textsc{B1} (Post-Norm decoder), \textsc{B2} (16-bit input encoding), \textsc{B3} (LayerNorm encoder).
\item \textbf{Refiner}: BFGS in place of Levenberg--Marquardt (Appendix~\ref{app:refiner_choice}).
\item \textbf{Decoding}: softmax sampling vs.\ beam search (Section~\ref{sec:ablations}).
\end{itemize}
Each variant perturbs exactly one axis and shares architecture, training data, optimizer, and 3M-step compute budget with the deployed \textsc{v23.0-120M}. The training-data, prior, and architecture variants additionally share their training batch with \textsc{B0}, a configuration-matched retrain of \textsc{v23.0-120M} used as the matched control for the training-dynamics plot (Figure~\ref{fig:training_curves_ablations_B}). Test-time accuracy comparisons (Figure~\ref{fig:training_data_ablations_120M_full_metrics}) use the deployed \textsc{v23.0-120M}.

Constant optimization in our inference pipeline (Section~\ref{sec:inference}) uses the Levenberg--Marquardt (LM) algorithm via \texttt{scipy.optimize.curve\_fit}. We re-run the FastSRB evaluation on the deployed 120M baseline with LM replaced by BFGS (\texttt{scipy.optimize.minimize}, \texttt{method=BFGS}), holding all other pipeline components (generation, decontamination, ranking) identical.

Figure~\ref{fig:refiner_choice} shows the full multi-metric comparison plotted against per-problem wall time. At every compute budget LM achieves higher recovery (fNRR/vNRR), higher fit success, lower validation FVU, lower expression-length ratios, and higher F1 score than BFGS; the gap is most visible at moderate compute (1--10\,s) and stabilizes around $\Delta\mathrm{vNRR}\approx +11$\,pp at large budgets. Matching BFGS's vNRR at $c=8192$ ($\approx 39\%$) places LM between $c=128$ and $c=512$, an order-of-magnitude reduction in compute at matched recovery.

\newpage
\subsection{Detailed Ablation Metrics}
\label{app:full_ablation_metrics}

\begin{figure}[!ht]
  \begin{center}
  \centering
  \includegraphics[width=0.78\columnwidth]{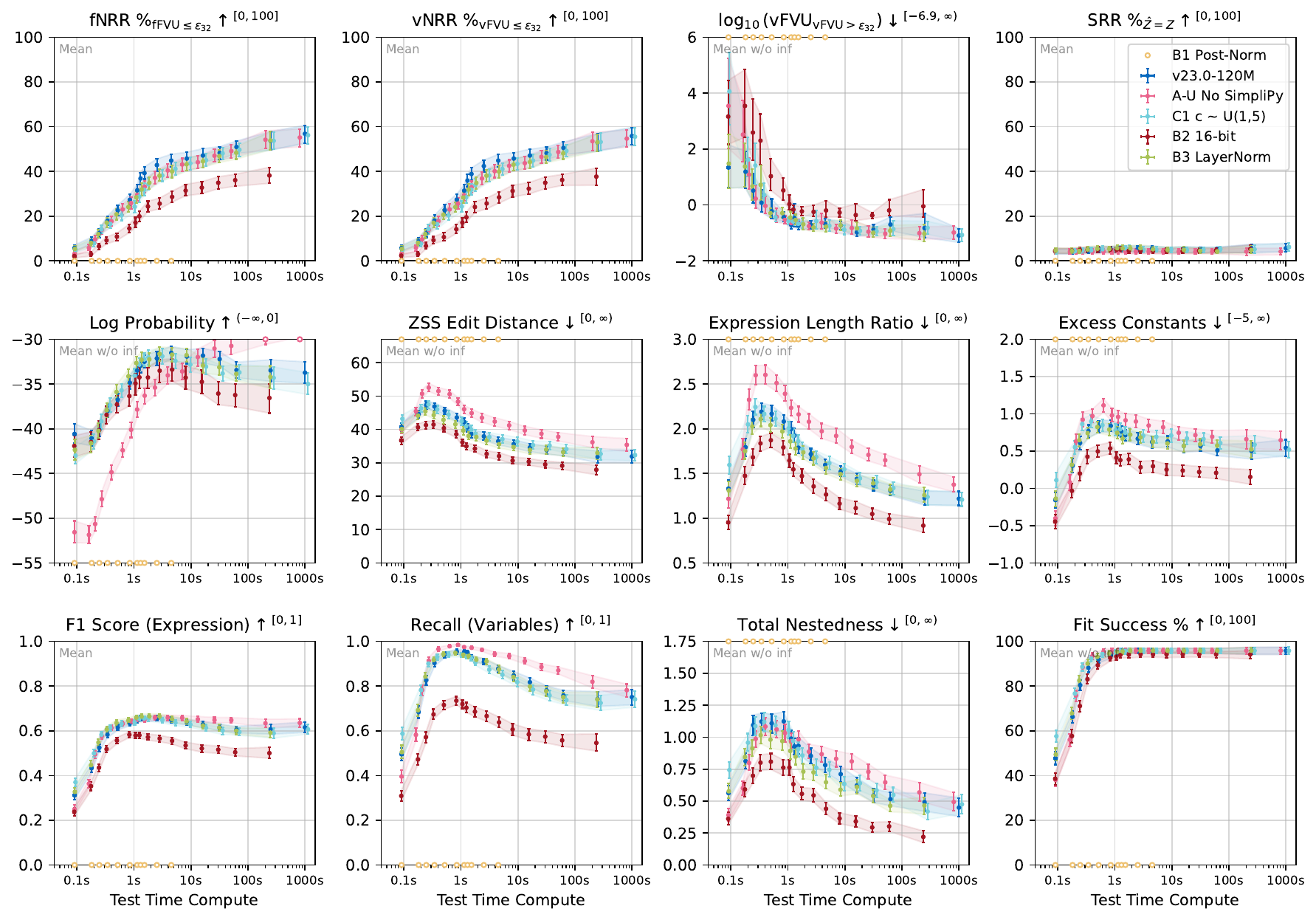}
  \caption{Time-normalized numerical and symbolic metrics for the five system-effect ablations (\textsc{A-U}, \textsc{B1}, \textsc{B2}, \textsc{B3}, \textsc{C1}) against the deployed \textsc{v23.0-120M} baseline. \textsc{A-U} and \textsc{B2} move recovery and parsimony metrics most. B1 fails completely due to unstable training.
  }
  \label{fig:training_data_ablations_120M_full_metrics}
  \end{center}
    \vspace{-20pt}
\end{figure}

\subsection{Refiner Choice: BFGS vs.\ Levenberg--Marquardt}
\label{app:refiner_choice}

\begin{figure}[!ht]
  \begin{center}
  \centering
  \includegraphics[width=0.78\columnwidth]{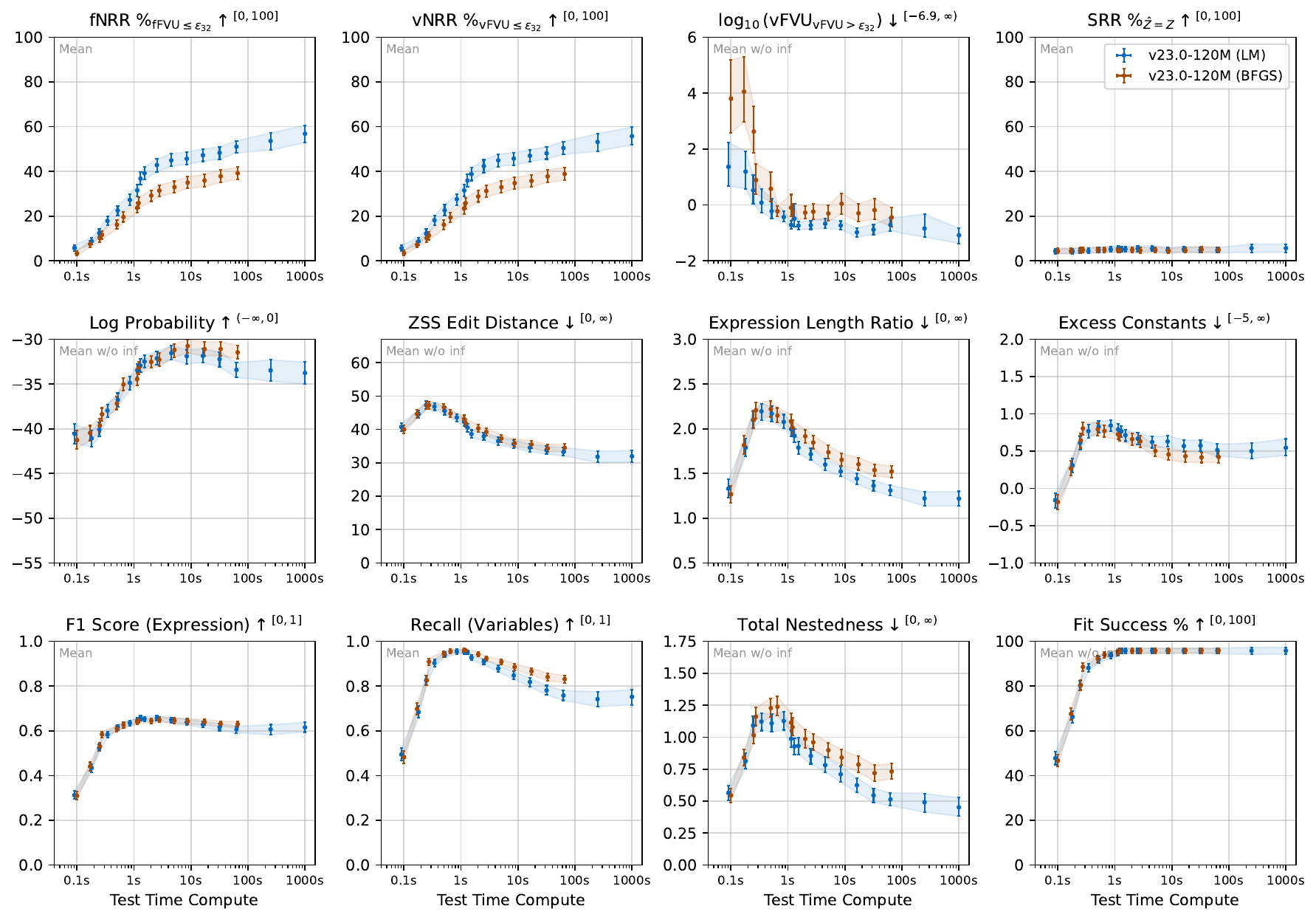}
  \caption{Time-normalized numerical and symbolic metrics for the BFGS vs.\ Levenberg--Marquardt (LM) refiner ablation. LM dominates BFGS at every budget ($\Delta\mathrm{vNRR}{\approx}+11$\,pp at large compute).}
  \label{fig:refiner_choice}
  \end{center}
\end{figure}

\newpage

\subsection{Constants Prior on the Gaussian Family}
\label{app:gaussian_heatmap}

On a Gaussian $f(x;\mu,\sigma) = \exp\!\bigl(-\tfrac{1}{2}(x-\mu)^2/\sigma^2\bigr)$ we sample $\mu \sim \mathcal{U}(-15, 15)$ and $\sigma \sim \mathrm{LogUniform}(10^{-3}, 10^{2})$, and fit $2{,}048$ instances per (model, budget) pair with \textsc{Flash-ANSR}.
At $512$ candidates both priors leave failure regions in the extreme-$|\mu|$ low-$\sigma$ corners ($|\mu| \approx 12$, $\sigma \approx 0.02$), where the Gaussian peak occupies a needle-thin window in the data: the broader $\mathcal{N}(0,5)$ prior fails roughly symmetrically in $\mu$, whereas the narrow $\mathcal{U}(1,5)$ prior of \citet{biggio2021neural} fails more on the negative-$\mu$ side, coinciding with its skewed prior.
At $8{,}192$ candidates both priors converge across most of the $(\mu, \sigma)$ plane.

\begin{figure}[ht]
  \begin{center}
    \centerline{\includegraphics[width=0.5\columnwidth]{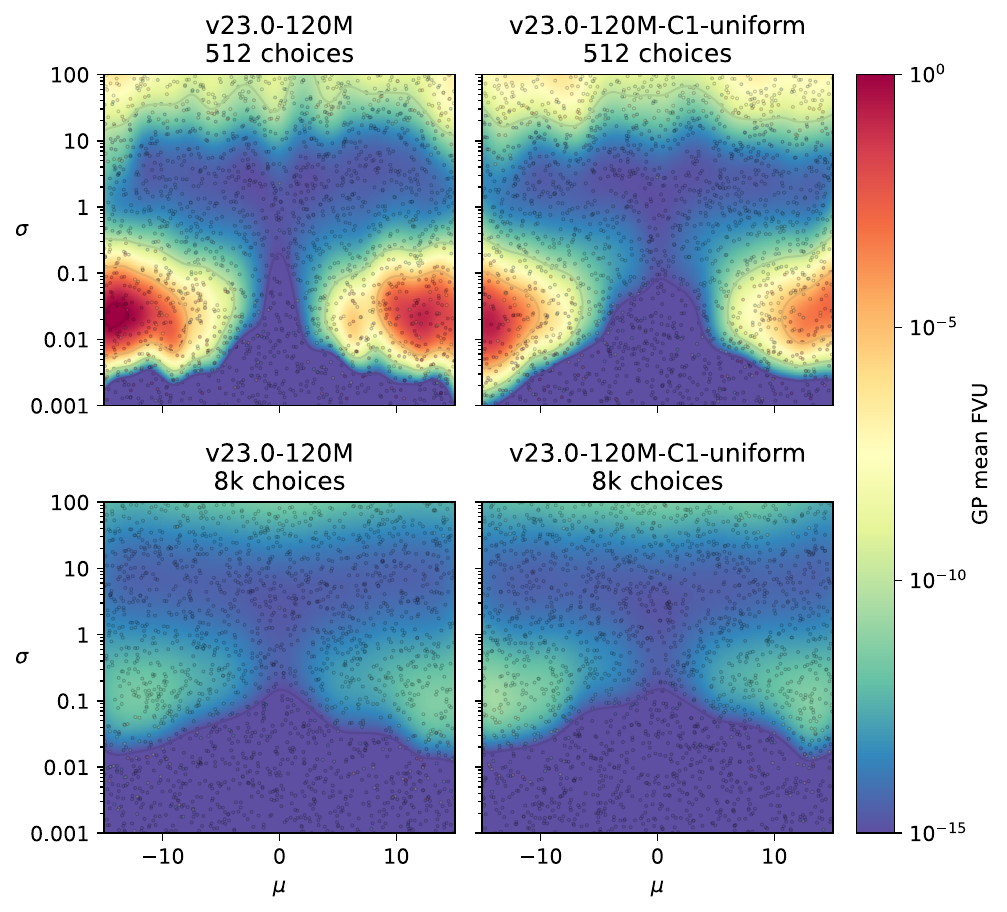}}
    \caption{Constants-prior comparison on the Gaussian family. Each panel shows per-instance $\log_{10}\mathrm{FVU}$ for $2{,}048$ random $(\mu, \sigma)$ draws (heatmap: Mat\'{e}rn-kernel Gaussian-process mean over $(\mu, \log\sigma)$; dots: raw samples coloured by their own FVU).
      \textbf{Top row:} $512$ candidate skeletons. \textbf{Bottom row:} $8{,}192$.
      \textbf{Left:} broader $\mathcal{N}(0, 5)$ prior used in this work. \textbf{Right:} narrow $\mathcal{U}(1, 5)$ prior of \citet{biggio2021neural}.
    }
    \label{fig:constants_variation_paper}
  \end{center}
\end{figure}
}

\newpage

\section{SimpliPy Rules}

\begin{table*}[!ht]
\caption{
  Random subset of the simplification rules from the \textsc{SimpliPy} engine organized into pattern-replacement pairs.
  The shown expressions are one-dimensional, hence we use $x$ as the variable token.
  The constant placeholder $\diamond$ represents an arbitrary finite constant.
  Infinities are carried through the simplification process to enable simplifications such as $\exp(-\frac{1}{x - x}) \xrightarrow{\text{Cancel}} \exp(-\frac{1}{0}) \xrightarrow{\text{Rule 108}} \exp(-\infty) \xrightarrow{\text{Rule 37}} 0$.
}
\label{tab:simplipy_rules_4col}
\vskip 0.15in
\begin{center}
\begin{small}
\renewcommand{\arraystretch}{1.8} 
\begin{tabular}{rcl | rcl}
\toprule
\textsc{Pattern} & $\to$ & \textsc{Replacement} & \textsc{Pattern} & $\to$ & \textsc{Replacement} \\
\midrule
$\left(\frac{-\infty}{4}\right)^{\sqrt[3]{\pi}}$ & $\to$ & $\infty$ & $\frac{4 x}{\pi}$ & $\to$ & $\diamond \cdot x$ \\
$(-\infty)^{x} \cdot (e \cdot x)$ & $\to$ & $(-\infty)^{x}$ & $(x \cdot \infty) + \frac{x}{\text{NaN}}$ & $\to$ & $\text{NaN}$ \\
$\frac{1/\pi}{0/(-\infty)}$ & $\to$ & $\infty$ & $\frac{x + (-\infty)}{\sinh(\pi)}$ & $\to$ & $-\infty$ \\
$(\infty \cdot 1) \cdot \frac{\pi}{-1}$ & $\to$ & $-\infty$ & $\frac{e}{-1} \cdot (-\infty)^\infty$ & $\to$ & $-\infty$ \\
$\sinh(\pi)^{x - \infty}$ & $\to$ & $0$ & $(x \cdot 0)^{-1/e}$ & $\to$ & $\infty$ \\
$\arcsin(0)^{\sqrt[3]{\pi}}$ & $\to$ & $0$ & $(\pi \cdot \pi) \cdot 2(-\infty)$ & $\to$ & $-\infty$ \\
$(-1 - \pi) \cdot (\infty \cdot \pi)$ & $\to$ & $-\infty$ & $\sinh(-\infty)^{\arctan(e)}$ & $\to$ & $\infty$ \\
$\frac{x + \infty}{5(-1)}$ & $\to$ & $-\infty$ & $(1/2)^{x/3}$ & $\to$ & $\diamond^{x}$ \\
$\frac{\infty^\pi}{\pi + (-1)}$ & $\to$ & $\infty$ & $\arcsin(-1) \cdot \infty^4$ & $\to$ & $-\infty$ \\
$\frac{\sqrt{-\infty}}{4\pi}$ & $\to$ & $\infty$ & $(\pi^\pi)^{\infty^3}$ & $\to$ & $\infty$ \\
\bottomrule
\end{tabular}
\end{small}
\end{center}
\vskip -0.1in
\end{table*}



\section{Memory Requirements}
\label{app:memory}

\camready{The pre-discovered rewrite table comes at a fixed memory cost.
\textsc{SimpliPy} occupies $\sim\!412$~MB of resident memory versus $\sim\!58$~MB for \textsc{SymPy}\footnote{Python 3.11, $L_{\max}=7$ rule set, RSS measured immediately after the engine class is loaded and the rule table is mapped into memory.}, a delta of $\sim\!350$~MB that does not grow with the size of the input expression and represents a small fraction of the total memory budget of a training run ($\geq\!16$~GB VRAM, CPU resident sets of the data loader, the transformer model, and per-worker simulation state).
We observe no simplification-induced memory growth during training or inference, including for expressions with up to $D = 17$ input variables and $|\bm{\tau}| \leq 35$ symbols (Appendix~\ref{app:data_statistics}).}

\camready{For amortized SR the binding constraint is throughput rather than memory.
At 100M expressions per training run with 8-thread parallel simplification (8T, matching our cluster training setup), $\sim\!1.7$~ms per expression with \textsc{SimpliPy} versus $\sim\!210$~ms with \textsc{SymPy} (mean, Section~\ref{sec:symbolic_simplification_efficiency}) translates to $\sim\!6$~hours versus $\sim\!30$~days of wall-clock simplification.
}

\newpage

\section{Per-Expression Results}

\begin{figure}[ht]
  \vskip 0.2in
  \begin{center}
    \centerline{\includegraphics[width=0.9\columnwidth]{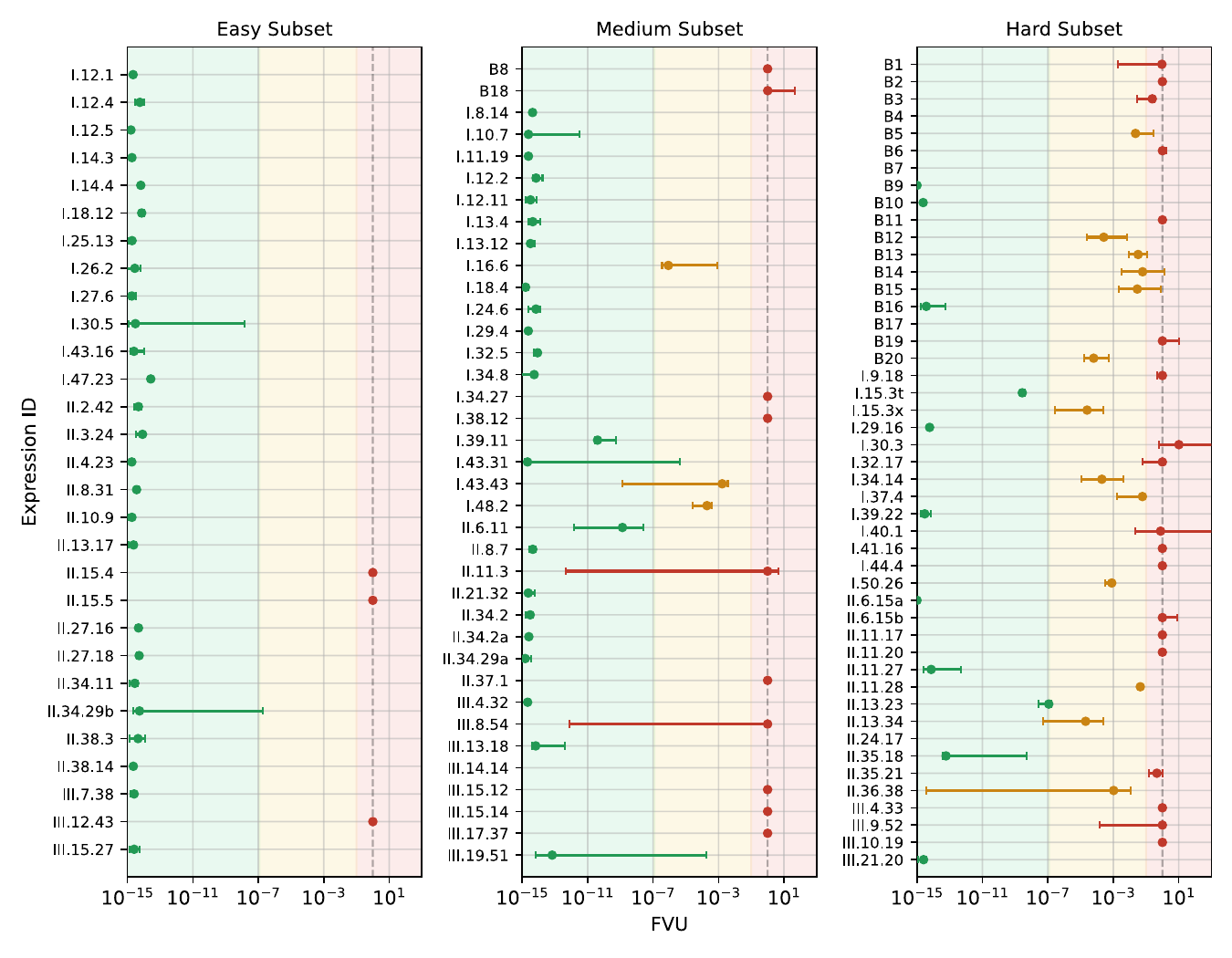}}
    \caption{
        Median and 95\% confidence intervals of the FVU with v23.0-120M, $\gamma = 0.05$, 8 optimizer restarts, 256k samples $\approx$ 1000s across individual expressions in the \textsc{FastSRB} benchmark stratified by difficulty level following \cite{matsubara2024rethinkingsymbolicregressiondatasets}.
        All but three ``easy'' expressions are solved perfectly (left, green dots). In both ``medium'' and ``hard'' categories, the share of approximate fits (orange) and failures (red, clustered around the gray dashed line at $\text{FVU} \approx 1$, i.e.\ no better than the mean predictor) grows with difficulty.
    }
    \label{fig:fastsrb_expression_fvu_by_difficulty}
  \end{center}
\end{figure}

\newpage

\section{Token Embeddings}

\begin{figure}[ht]
  \vskip 0.2in
  \begin{center}
    \begin{subfigure}{0.48\columnwidth}
      \centering
      \includegraphics[width=\linewidth]{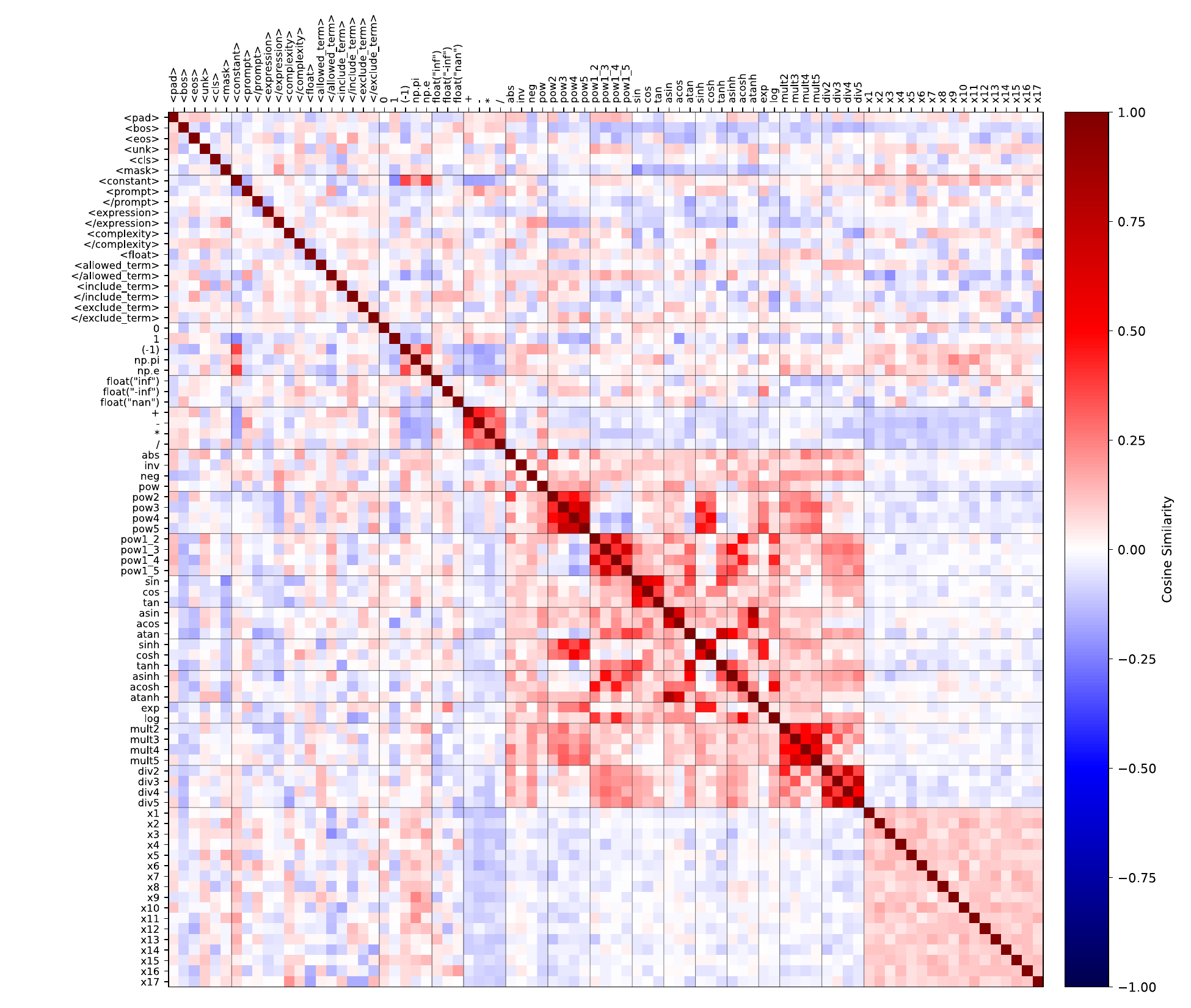}
      \caption{3M}
    \end{subfigure}\hfill
    \begin{subfigure}{0.48\columnwidth}
      \centering
      \includegraphics[width=\linewidth]{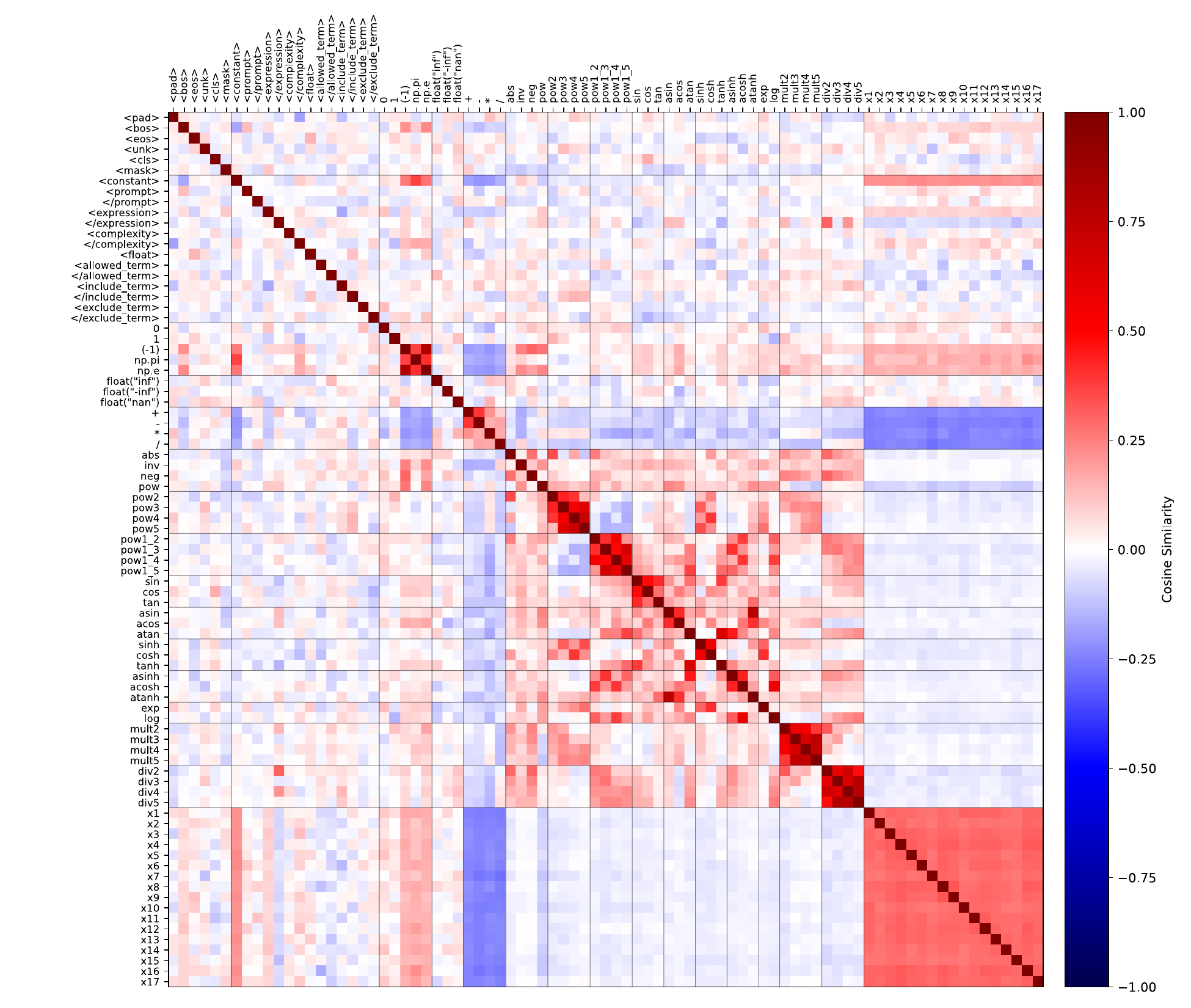}
      \caption{20M}
    \end{subfigure}

    \vskip 0.15in

    \begin{subfigure}{0.48\columnwidth}
      \centering
      \includegraphics[width=\linewidth]{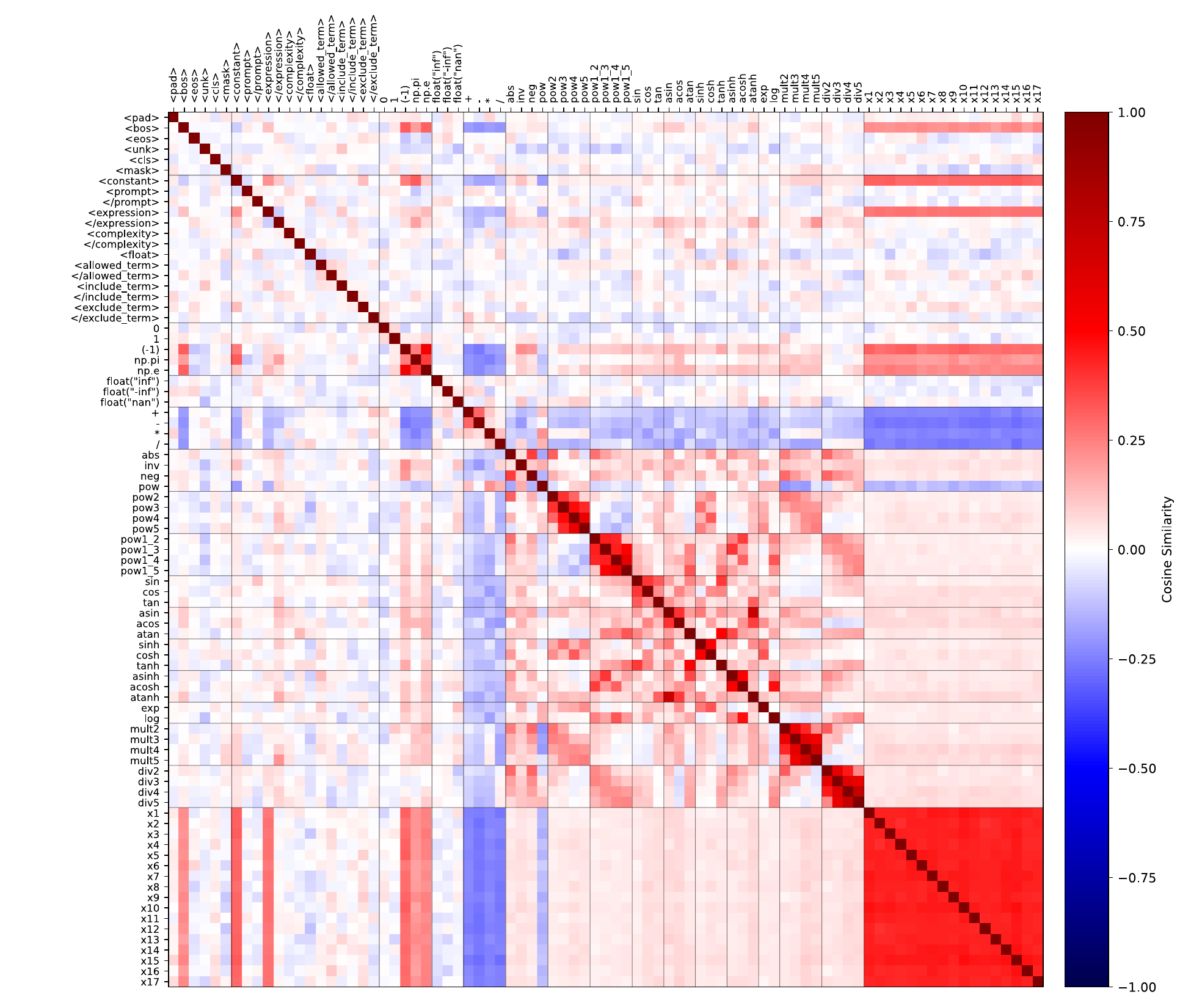}
      \caption{120M}
    \end{subfigure}\hfill
    \begin{subfigure}{0.48\columnwidth}
      \centering
      \includegraphics[width=\linewidth]{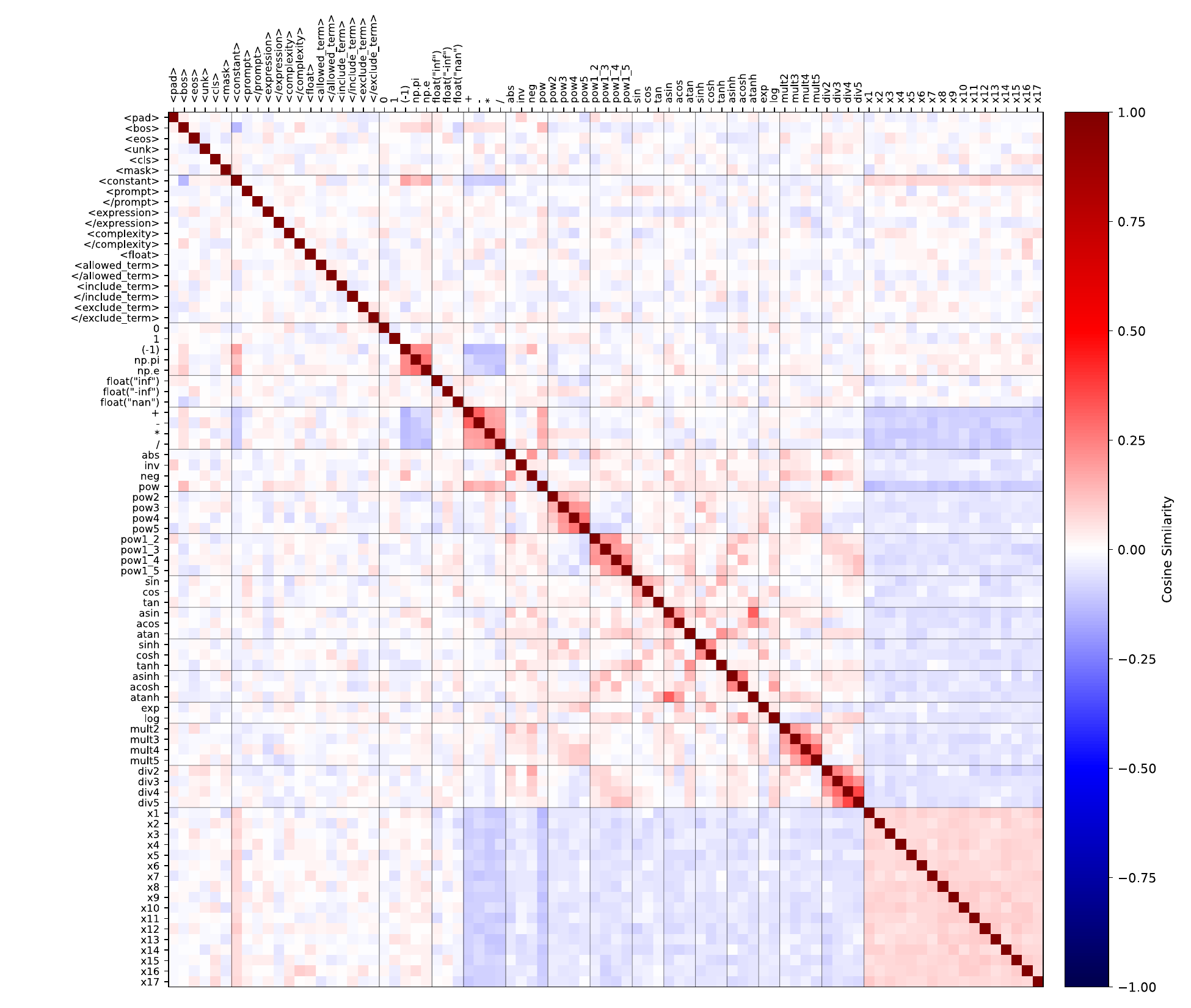}
      \caption{1B}
    \end{subfigure}

    \caption{Pairwise token embedding similarity for the Flash-ANSR model family (v23.0) across model sizes. Across all scales, we observe high similarity of semantically related operators (e.g., binary operators, trigonometric functions, exponentials and powers) and variable tokens, and low similarity between different types of tokens (e.g., operators vs.\ variables). The overall strength of these relationships varies across model sizes with no discernible pattern.}
    \label{fig:operator_embeddings_grid}
  \end{center}
\end{figure}


\end{document}